\documentclass[11pt]{article}

\usepackage{acl}

\usepackage{times}
\usepackage{latexsym}

\usepackage[T1]{fontenc}
\usepackage[utf8]{inputenc}

\usepackage{microtype}

\usepackage{amsmath}
\usepackage{enumitem}
\usepackage{graphicx}
\usepackage{subcaption}
\usepackage{wrapfig}
\usepackage{adjustbox}
\usepackage{array}
\usepackage{booktabs}
\usepackage{amsfonts}
\usepackage{nicefrac}
\usepackage{xcolor}
\usepackage{url}
\usepackage{hyperref}
\usepackage{multirow}
\usepackage{booktabs}  
\usepackage[table]{xcolor} 
\usepackage{fontawesome5} 

\definecolor{posgreen}{HTML}{008000} 
\definecolor{negred}{HTML}{C00000}   

\definecolor{headerbg}{HTML}{F2F4F7}    
\definecolor{defaultbg}{HTML}{EBF3FB}   
\definecolor{subtext}{HTML}{666666}     
\definecolor{subgray}{HTML}{666666}

\title{Improving Parameter Utilization by Sharing Neural Experts Across Layers in Transformers}

\author{
  \textbf{Dian Jiao}\textsuperscript{*},
  \textbf{Jiaxin Duan}\textsuperscript{*},
  \textbf{Shuai Zhao},
  \textbf{Jiabing Leng},
  \textbf{Yiran Zhang},
  \textbf{Feng Huang} \\  %
  China Electronics Cloud Technology Co., Ltd. \\
  \texttt{\{jiaodian,duanjiaxin,zhaoshuai\}@cestc.cn} \\
  \texttt{\{lengjiabing,zhangyiran,huangfeng01\}@cestc.cn} \\
  \small\faGithub\ \url{https://github.com/CESTC-REAL/Self-MoE}
}

\begin{document}
\maketitle

\renewcommand{\thefootnote}{\fnsymbol{footnote}} 

\footnotetext[1]{Equal contributions.} 


\begin{abstract}
Transformer-based large language models often suffer from inter-layer parameter redundancy, where functional transformations are redundantly learned across network depths. We propose CS-MoE, a novel Transformer architecture featuring cross-layer expert sharing to address this inefficiency. 
Deviating from the widely used Mixture-of-Experts (MoE) architecture that terminates each Transformer block with layer-isolated experts, CS-MoE combines layer-independent experts with concurrent access to a centralized, globally shared expert pool. This \textit{Global Experts Sharing} mechanism enables elastic control over token-level parameter activation and computational consumption (FLOPs). Experiments demonstrate that CS-MoE achieves lower perplexity than equal-scale dense Transformers while activating only 55\% of parameters. Furthermore, its performance scales monotonically with an increased number of activated experts and approaches MoE counterparts that consume more FLOPs by expanding the shared pool without increasing FLOPs. CS-MoE also establishes a flexible Pareto frontier between computational cost and model capacity, offering an efficient alternative for computation-constrained environments.
\end{abstract}

\section{Introduction}

The proliferation of Transformer-based architectures has catalyzed the success of Large Language Models (LLMs), but scaling toward trillion-parameter regimes yields diminishing marginal returns in efficiency~\cite{dai2024deepseekmoe,muennighoff2025olmoe}. Empirical inquiries into weight pruning, model compression~\cite{shazeer2017outrageously,lepikhin2020gshard}, and activation dynamics suggest that dense networks suffer from significant activation redundancy, wherein a substantial portion of Feed-Forward Network (FFN) parameters contributes negligibly to the final output for any given input token~\cite{sparseswaps2025}. This latent underutilization leads to a suboptimal ratio of information density to computational expenditure during inference.

\begin{figure}[!t]
\centering
\includegraphics[width=1.0\linewidth]{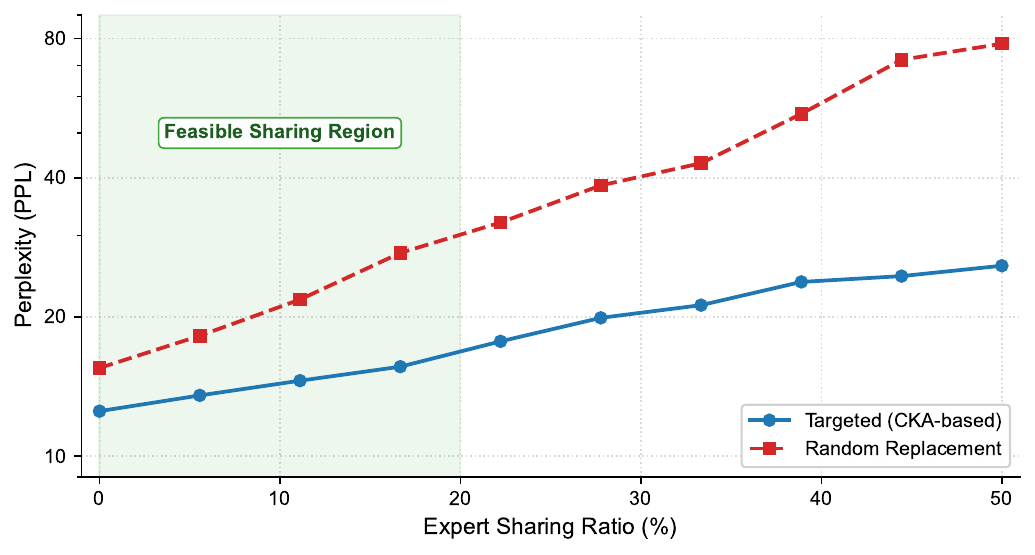}
\caption{
PPL of MoE models with random expert replacement and deliberate expert replacement based on CKA analysis.
}
\label{fig:0}
\end{figure}

The Mixture-of-Experts (MoE) paradigm has emerged as a standard solution, substituting dense layers with sparse experts and selectively activating a subset (Top-$K$) per token to decouple total model capacity from active computation (FLOPs)~\cite{Yun2024TowardIM,jiang2024mixtralexperts}. However, conventional MoE designs rely on layer-isolated experts, introducing parameter redundancy by preventing the network from exploiting functional recurrence across different depths~\cite{jin2024moeacceleratingmixtureofexpertsmethods,wen2025routeexpertssequencetoken,yang-etal-2024-xmoe}. To investigate this, we conduct a pilot study replacing deep-layer experts with functionally equivalent shallow counterparts identified via Centered Kernel Alignment (CKA) (details in Appendix~\ref{sec:appendix-pilot}). As illustrated in Figure~\ref{fig:0}, this CKA-based expert substitution incurs only moderate perplexity degradation and substantially outperforms random replacement, indicating that experts are largely interchangeable across depths.

Based on these insights, we propose CS-MoE, a novel Transformer architecture that bypasses layer isolation via a \textit{Global Experts Sharing} mechanism. First, a \textit{Fixed Path} comprising a small set of \textit{Independent Experts} is unconditionally activated at each layer to capture depth-specific features such as hierarchical syntax. Additionally, a \textit{Dynamic Path} grants all layers concurrent access to a global expert pool. By employing a layer-specific router to select proportional experts from this global repository, CS-MoE preserves the hierarchical integrity of the Transformer while maximizing parameter reutilization.

In experiments, we validate the efficacy of CS-MoE across model scales ranging from 0.6B to 12B parameters. Extensive results demonstrate its threefold advantages: 1) CS-MoE achieves superior parameter efficiency, outperforming dense baselines in perplexity while requiring only 55\% active parameters. 2) Its performance scales monotonically with the expert activation count ($K$), leveraging the capacity of the shared pool; 3) The architecture offers a flexible Pareto frontier, providing a highly efficient alternative in FLOP-constrained environments equivalent to the performance of standard MoE by scaling the shared experts pool.

\section{Methodology}

\subsection{Formalization and Topology}

We formalize the Transformer Feed-Forward Network through a generalized Mixture-of-Experts (MoE) framework. In this view, a standard \textit{dense Transformer} represents a constrained instance of this topology with a single, active expert ($N=1, K=1$) per layer. A \textit{sparse MoE Transformer} scales parameter capacity via a larger pool of experts ($N > 1$) while restricting active computation to a small subset ($K \ll N$). 
\textit{CS-MoE} generalizes these paradigms into a dual-tier topology. It anchors a localized set of experts to specific layers to capture depth-dependent features, while providing a centralized, shared expert pool accessible across all layers. This design allows individual experts to participate in multiple stages of the network, bypassing traditional layer isolation to maximize parameter utility and computational efficiency.

\begin{figure}[!t]
\centering
\includegraphics[width=0.95\linewidth]{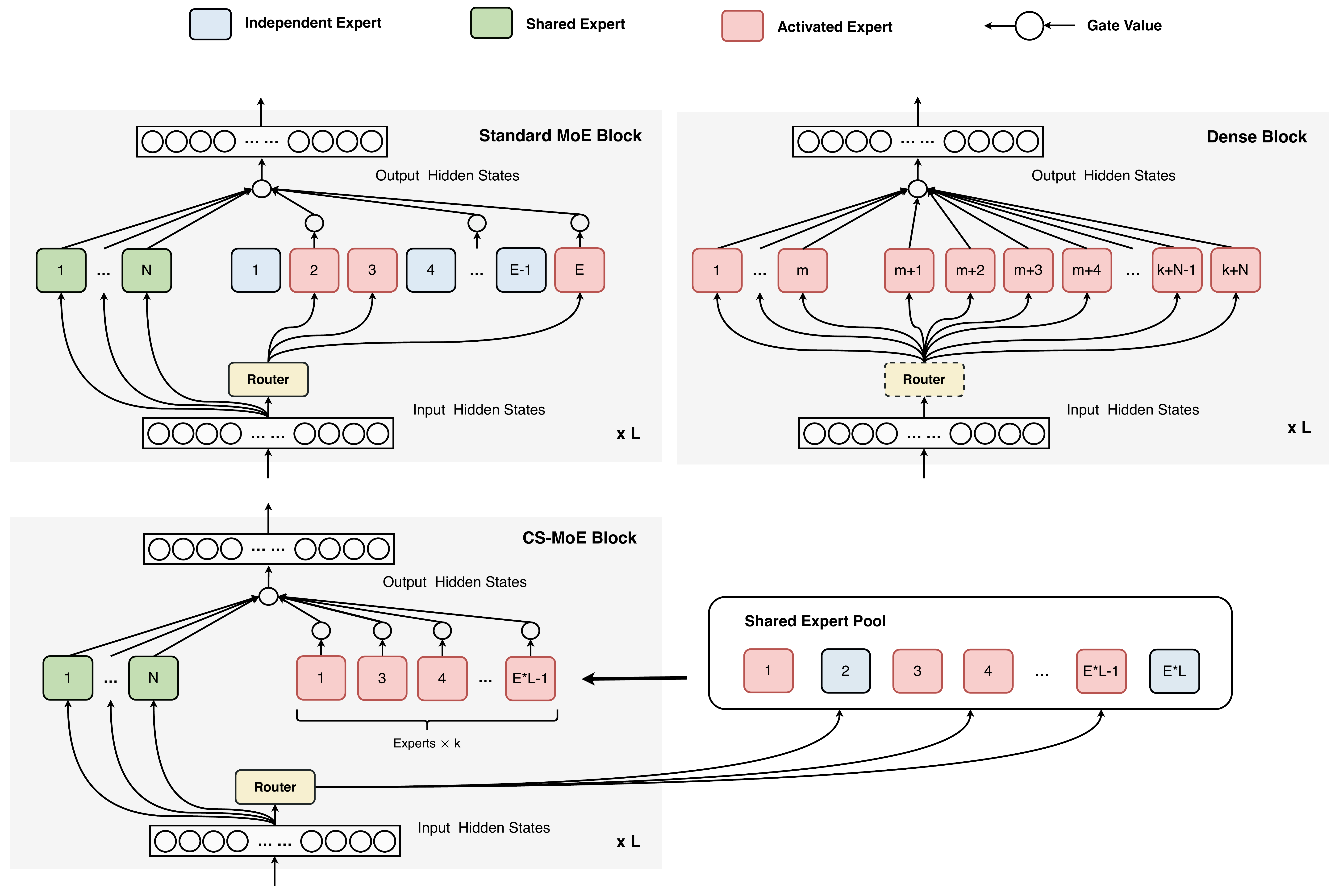}
\caption{
Overview of Transformer architectures: MoE (upper left), dense (upper right), and our CS-MoE (lower). 
The dashed box indicates a virtual router.
}
\label{fig:c}
\end{figure}

\subsection{CS-MoE Architecture}

As illustrated in Figure~\ref{fig:c}, CS-MoE restructures each Transformer layer $l \in \{1, \dots, L\}$ into a dual-path execution model: a \textbf{Fixed Path} consisting of layer-specific Independent Experts and a \textbf{Dynamic Path} drawing from a global Shared Expert Pool. The total expert set available to the network is partitioned as follows:
$$
\mathcal{E}_{total} = \left( \bigcup_{l=1}^{L} \mathcal{E}_{indep}^{(l)} \right) \cup \mathcal{E}_{shared}
$$
where $\mathcal{E}_{indep}^{(l)} = \{e_1^{(l)}, \dots, e_{N}^{(l)}\}$ contains $N$ independent experts private to layer $l$, and $\mathcal{E}_{shared} = \{s_1, \dots, s_M\}$ is a large, centralized repository of $M$ experts accessible by every layer in the Transformer stack. In this configuration, the execution logic for each layer follows two distinct principles:

\begin{itemize}[leftmargin=*]
\item \textbf{Routing-free Fixed Path:} Each layer $l$ activates its own $N$ \textit{Independent Experts} ($\mathcal{E}_{indep}^{(l)}$) unconditionally for every token. This ensures a baseline of depth-specific transformation capacity without the overhead of routing, effectively maintaining the reliability of dense architectures.
\item \textbf{Sparse Dynamic Path:} The routing mechanism in CS-MoE is exclusively restricted to the \textit{Shared Expert Pool}. For each token at layer $l$, a layer-specific router selects and activates $k$ experts from $\mathcal{E}_{shared}$. Crucially, because the shared pool is universal, a specific shared expert $s_j$ may be activated simultaneously across multiple layers for the same or different tokens, allowing for extreme longitudinal parameter reuse.
\end{itemize}

Consequently, the total computation of a single layer is composed of $N$ independent experts and $k$ routed shared experts, resulting in a total of $k+N$ activated experts per layer. This hybrid structure allows CS-MoE to simulate the capacity of an ultra-wide MoE while maintaining a lean physical parameter footprint.

\subsection{Hybrid Activation and Routing Strategy}

The execution of a CS-MoE layer is defined by the simultaneous processing of these two paths. Given a token representation $\mathbf{h}_l \in \mathbb{R}^d$, the routing mechanism manages only the dynamic path (the shared pool).

\paragraph{Shared Path Routing:} The layer-specific router $R_l$ computes scores $\mathbf{s}_l$ for the $M$ shared experts:
\[
\mathbf{s}_l = W_{r,l} \mathbf{h}_l
\]
where $W_{r,l} \in \mathbb{R}^{M \times d}$. We then apply a Top-$K$ gating function to select the most relevant shared experts:
\[
\mathbf{g}_l = \text{Softmax}(\text{TopK}(\mathbf{s}_l, K))
\]

\paragraph{Output Aggregation:} The final output of the layer, $\mathbf{o}_l$, is the sum of the dense independent experts and the sparsely activated shared experts:
\[
\mathbf{o}_l = \underbrace{\sum_{j=1}^{N_{indep}} \text{Expert}_{e,j}^{(l)}(\mathbf{h}_l)}_{\text{Fixed Path}} + \underbrace{\sum_{i \in \text{TopK}} g_{l,i} \cdot \text{Expert}_{s,i}(\mathbf{h}_l)}_{\text{Dynamic Path}}
\]

This design ensures that every token benefits from $N_{indep}$ specialized parameters tailored to the current layer's depth, while simultaneously drawing from $K$ experts in the global pool. By adjusting $K$ and $M$, the architecture can precisely tune the ratio of compute-to-parameters, effectively breaking the linear relationship between model depth and parameter count.

\subsection{Layer-wise Load Balancing}

To prevent the shared pool from suffering from expert collapse - where a few experts dominate the routing across all layers - we implement a layer-wise auxiliary load-balancing loss $\mathcal{L}_{aux}^{(l)}$. This loss is calculated specifically for the shared expert pool at each layer to ensure diverse utilization.

For a batch of $T$ tokens, the balancing loss is defined as:
\[
\mathcal{L}_{aux}^{(l)} = M \sum_{i=1}^{M} f_{l,i} \cdot P_{l,i}
\]
Where $f_{l,i}$ is the fraction of tokens in the batch that selected shared expert $i$ at layer $l$, and $P_{l,i}$ is the average routing probability assigned to that expert. The total objective function is:
\[
\mathcal{L}_{total} = \mathcal{L}_{LM} + \alpha \sum_{l=1}^{L} \mathcal{L}_{aux}^{(l)}
\]
where $\alpha$ is the balancing coefficient. This ensures that the shared pool remains a robust, multi-functional resource throughout the training process.

\begin{figure*}[!ht]
    \centering
    \begin{subfigure}[b]{0.23\textwidth}
        \centering
        \includegraphics[width=\textwidth]{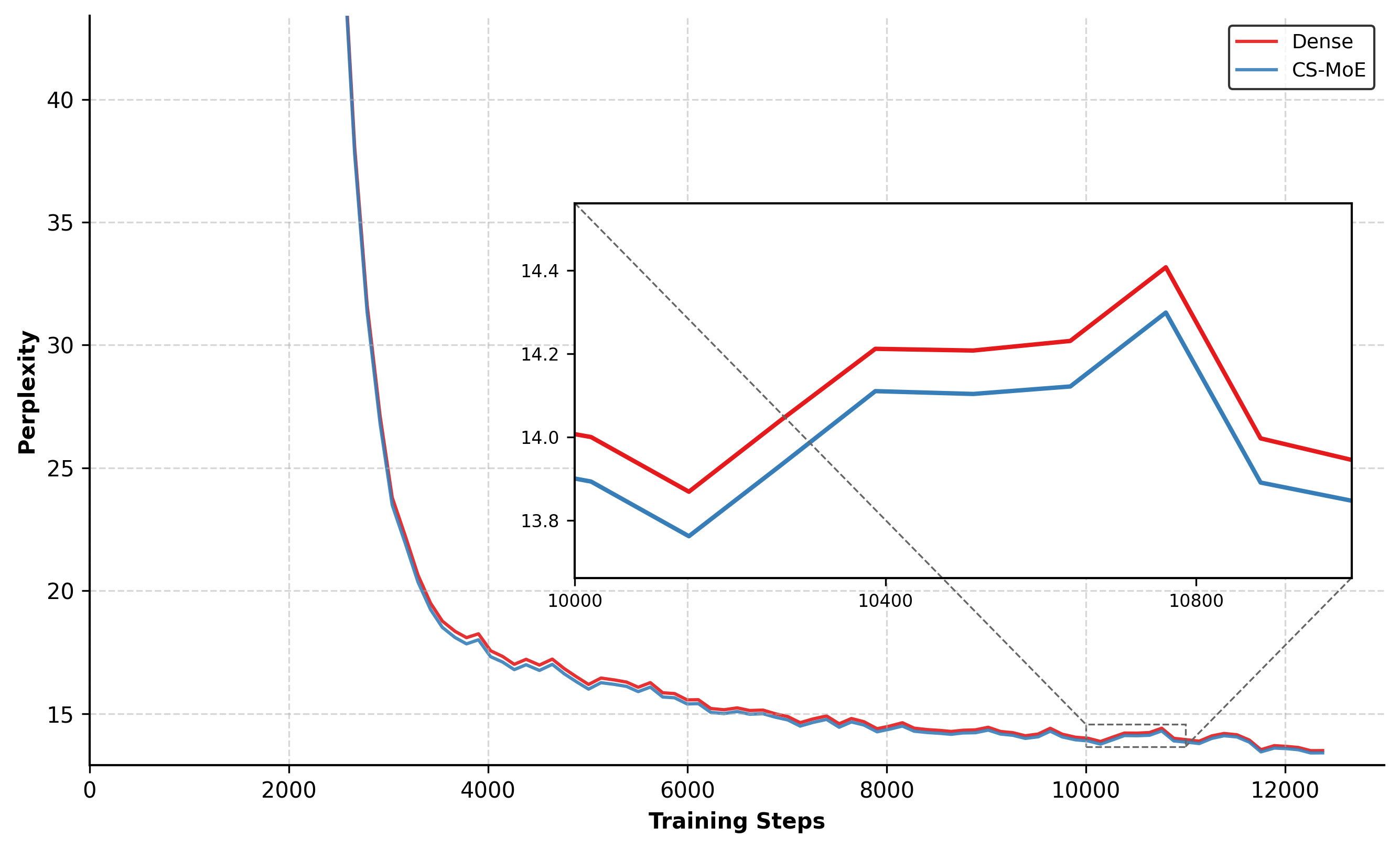}
        \caption{PPL of 0.6B Models.}
        \label{fig1:sub1}
    \end{subfigure}
    \hfill
    \begin{subfigure}[b]{0.23\textwidth}
        \centering
        \includegraphics[width=\textwidth]{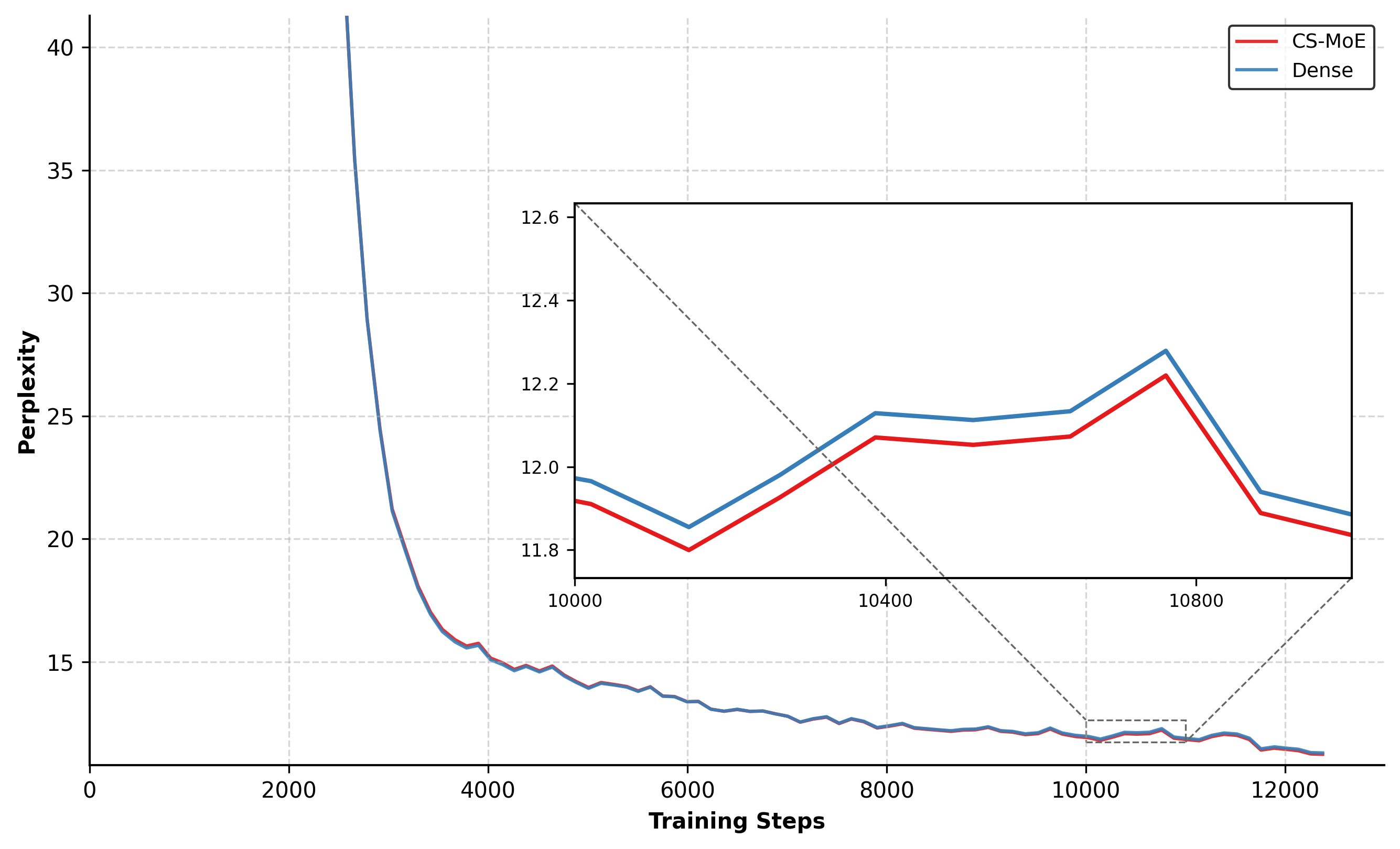}
        \caption{PPL of 1.7B Models.}
        \label{fig1:sub2}
    \end{subfigure}
    \hfill
    \begin{subfigure}[b]{0.23\textwidth}
        \centering
        \includegraphics[width=\textwidth]{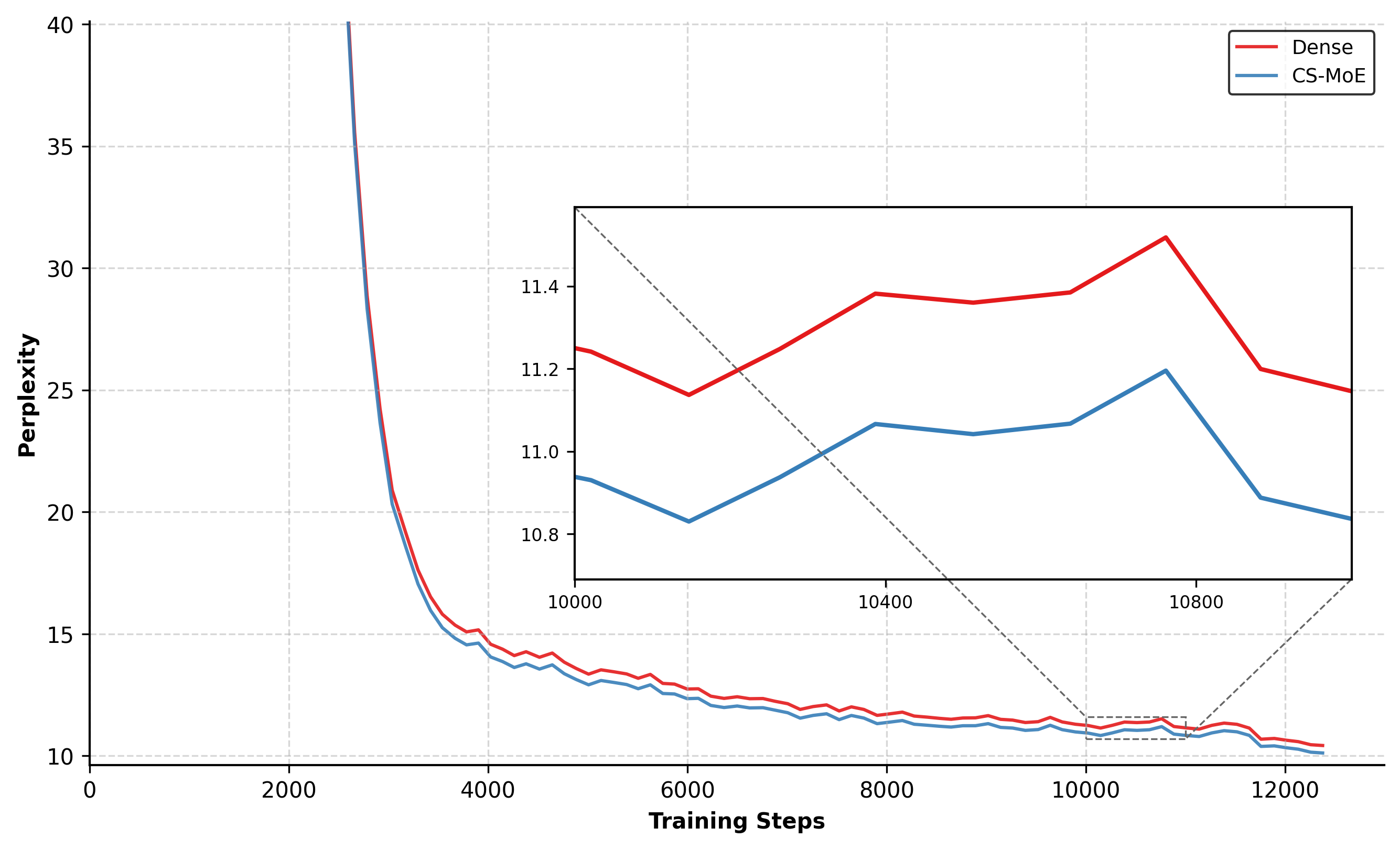}
        \caption{PPL of 4B Models.}
        \label{fig1:sub3}
    \end{subfigure}
    \hfill
    \begin{subfigure}[b]{0.23\textwidth}
        \centering
        \includegraphics[width=\textwidth]{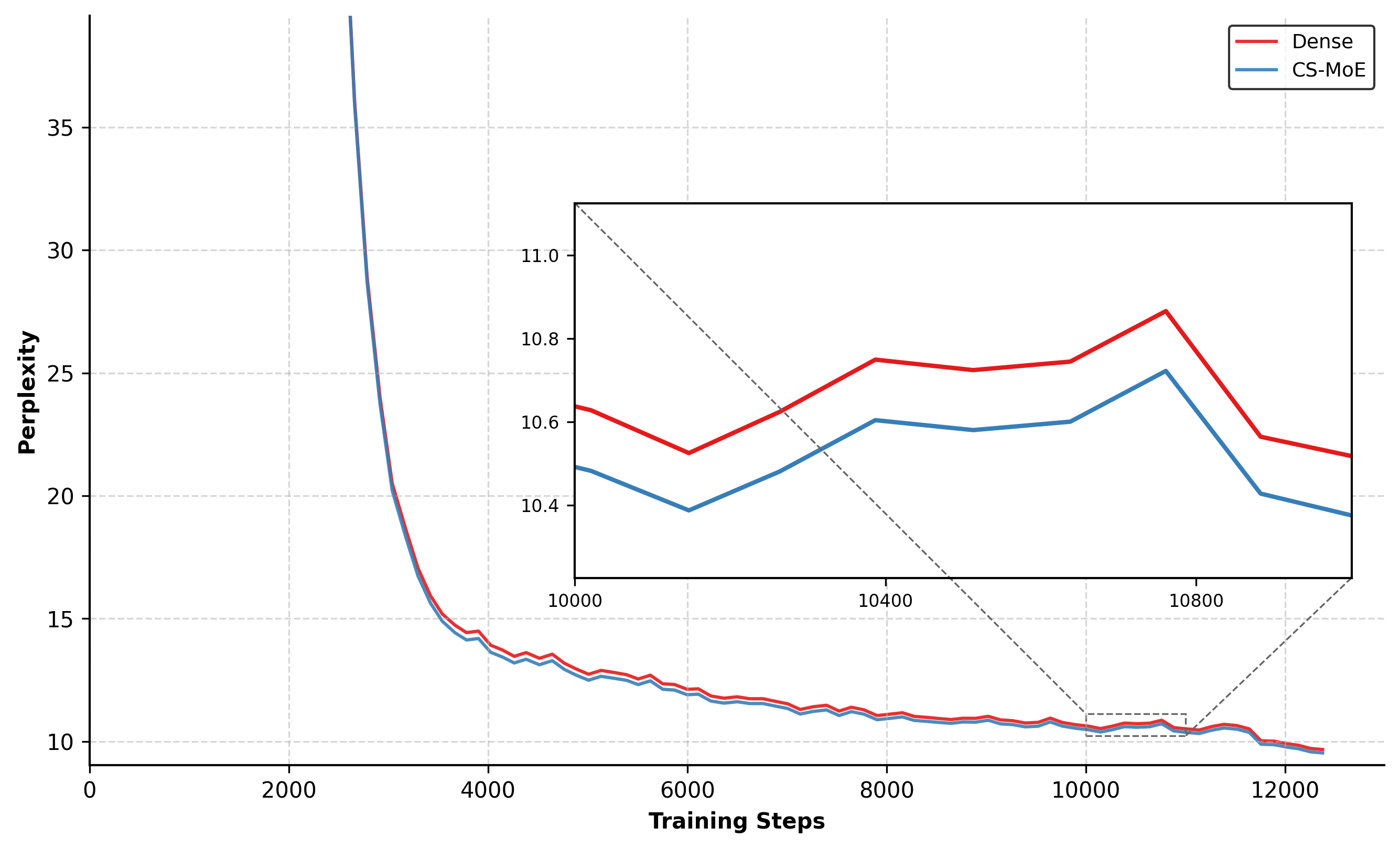}
        \caption{PPL of 8B Models.}
        \label{fig1:sub4}
    \end{subfigure}
    \caption{Comparison between CS-MoE and Dense models.}
    \label{fig:1}
\end{figure*}

\begin{figure*}[!ht]
    \centering
    \begin{subfigure}[b]{0.25\textwidth}
        \centering
        \includegraphics[width=\textwidth]{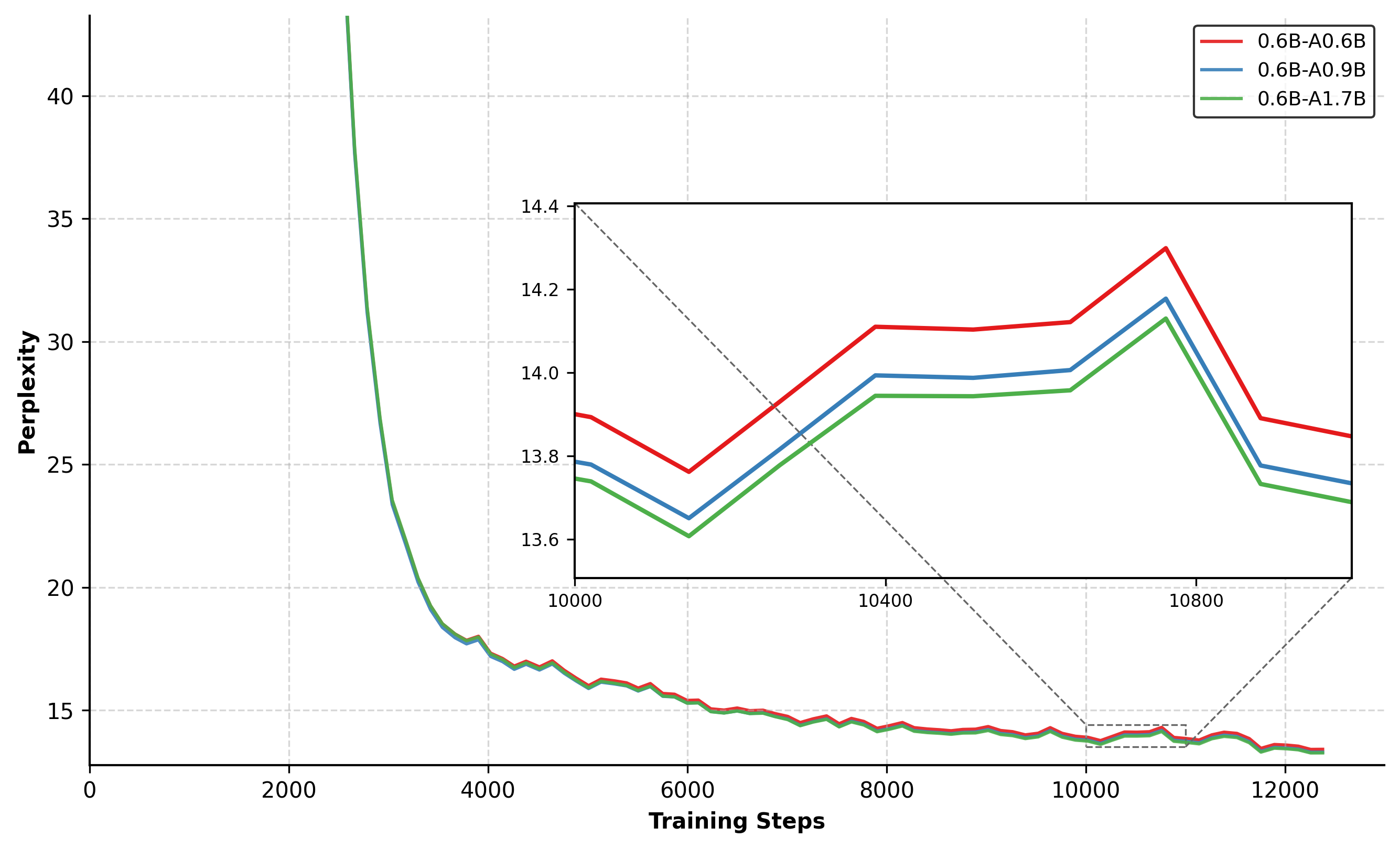}
        \caption{PPL of 0.6B-AxB Models.}
        \label{fig2:sub1}
    \end{subfigure}
    \begin{subfigure}[b]{0.25\textwidth}
        \centering
        \includegraphics[width=\textwidth]{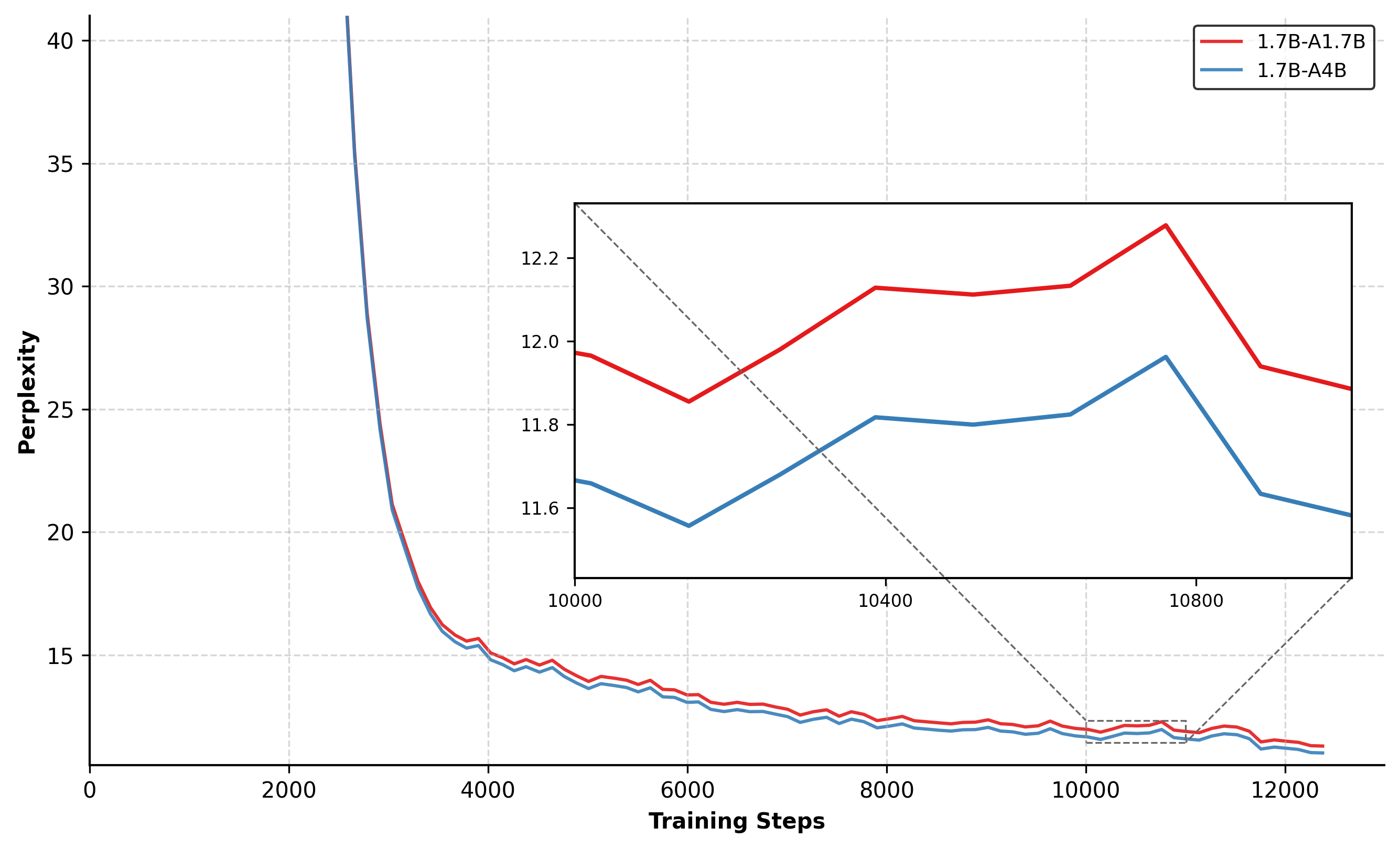}
        \caption{PPL of 1.7B-AxB Models.}
        \label{fig2:sub2}
    \end{subfigure}
    \begin{subfigure}[b]{0.25\textwidth}
        \centering
        \includegraphics[width=\textwidth]{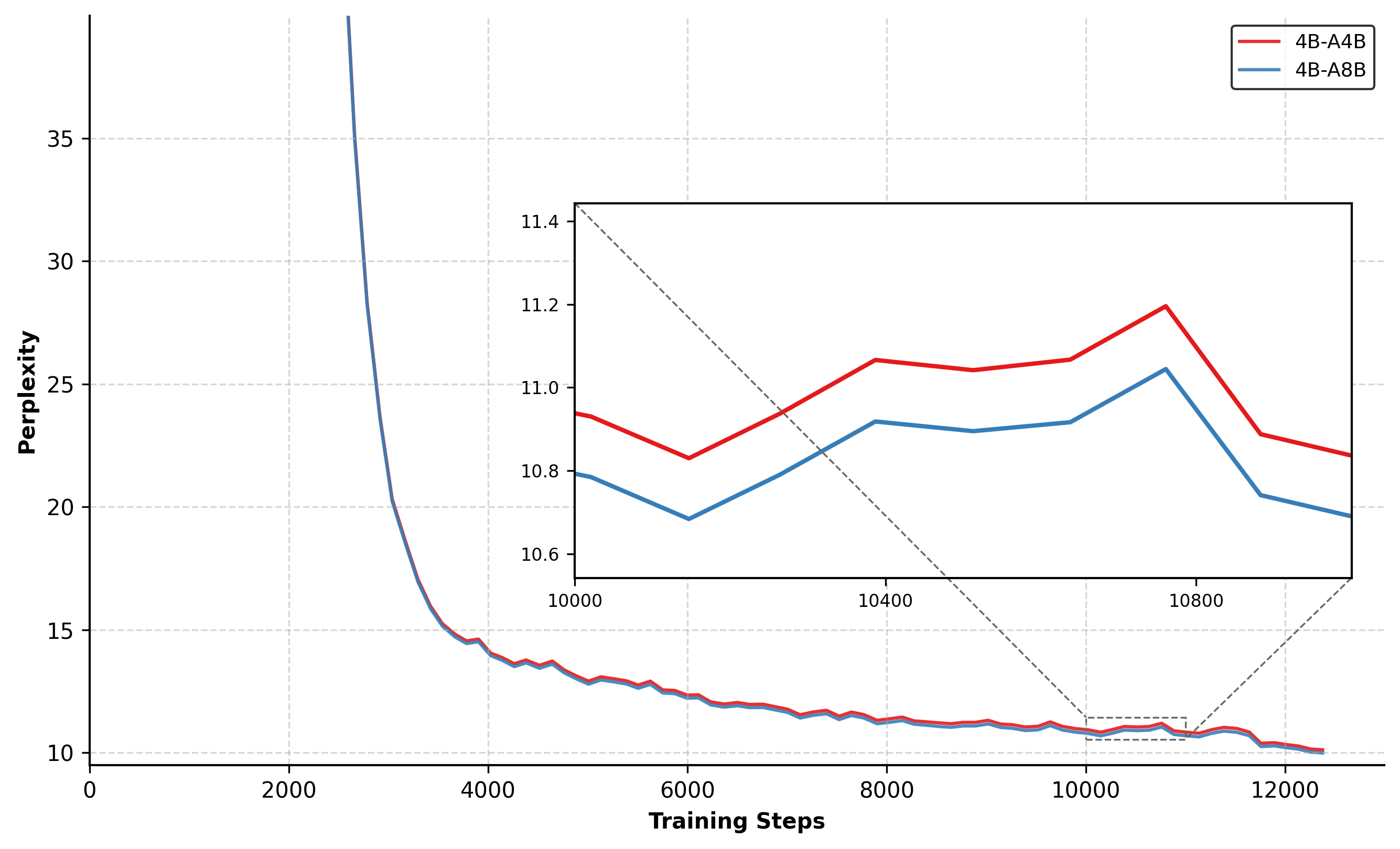}
        \caption{PPL of 4B-AxB Models.}
        \label{fig2:sub3}
    \end{subfigure}
    \caption{Comparison among CS-MoE models with varying activations.}
    \label{fig:2}
\end{figure*}

\begin{figure}[!ht]
    \centering
    \begin{subfigure}[b]{0.48\linewidth}
        \centering
        \includegraphics[width=\linewidth]{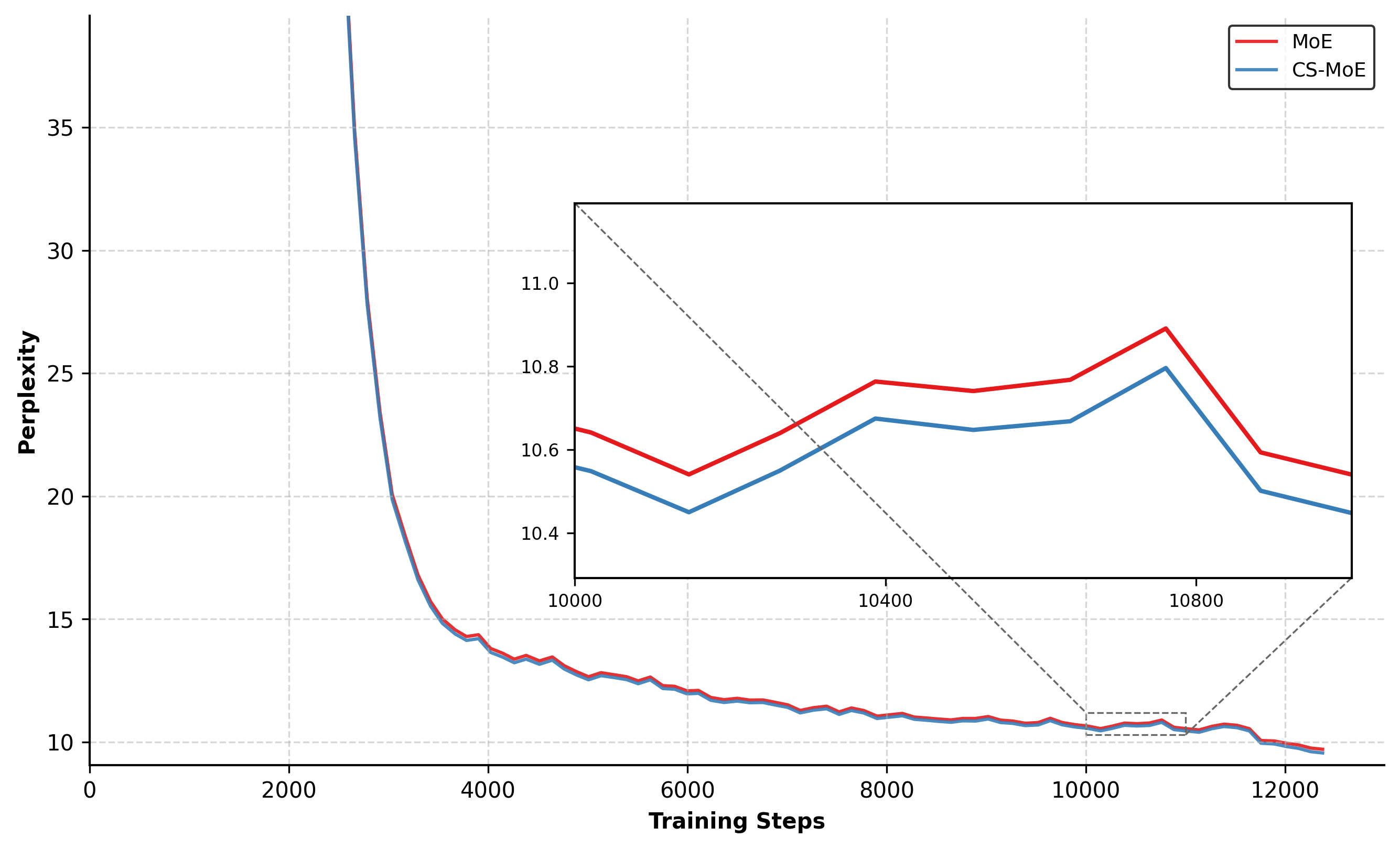}
        \caption{PPL of 8B-A4B Models.}
        \label{fig3:sub1}
    \end{subfigure}
    \hfill
    \begin{subfigure}[b]{0.48\linewidth}
        \centering
        \includegraphics[width=\linewidth]{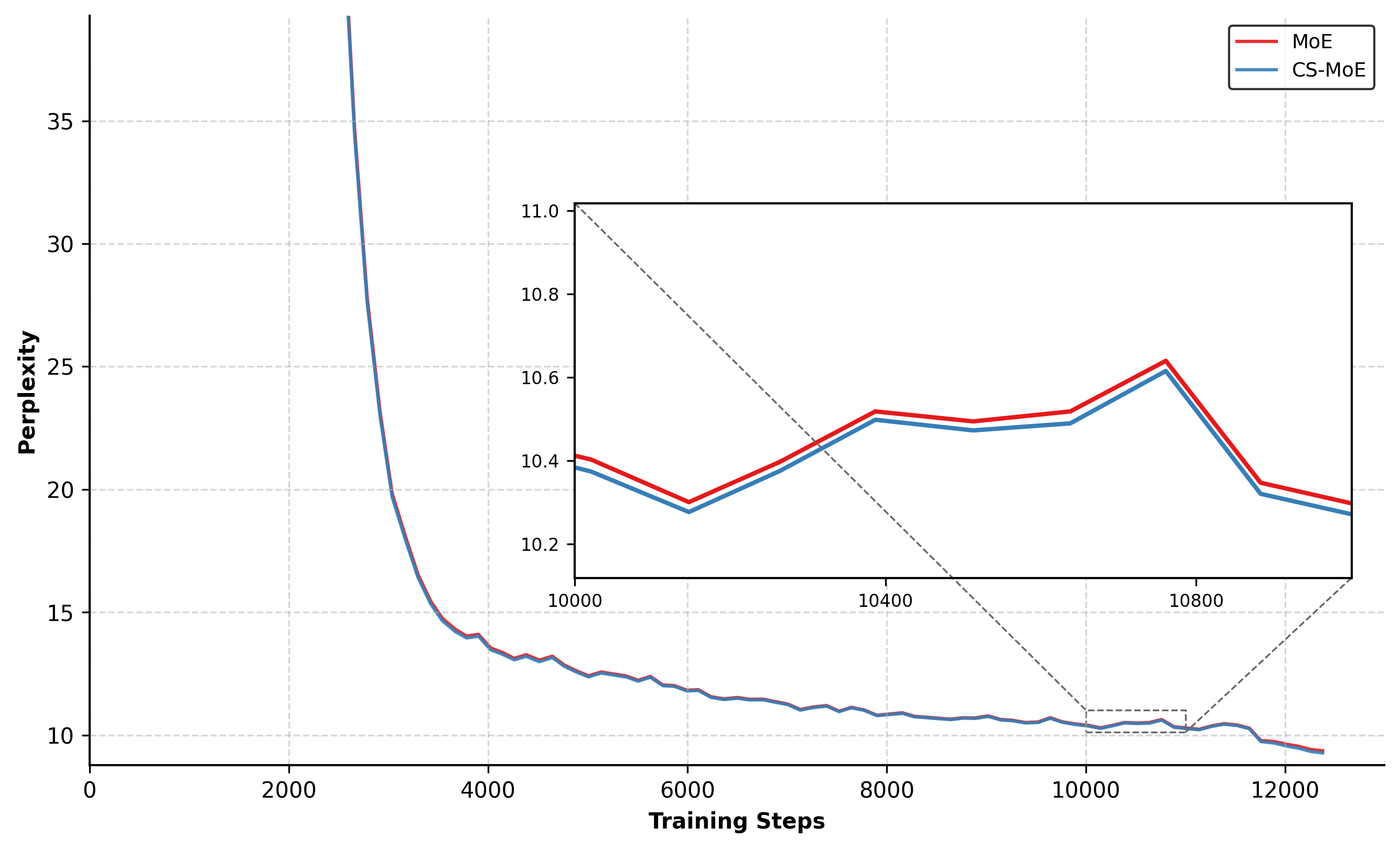}
        \caption{PPL of 12B-A4B Models.}
        \label{fig3:sub2}
    \end{subfigure}
    \caption{Comparison between CS-MoE and Sparse MoE models with equal activations.}
    \label{fig:3}
\end{figure}

\begin{figure}[!ht]
\centering
\includegraphics[width=0.48\textwidth]{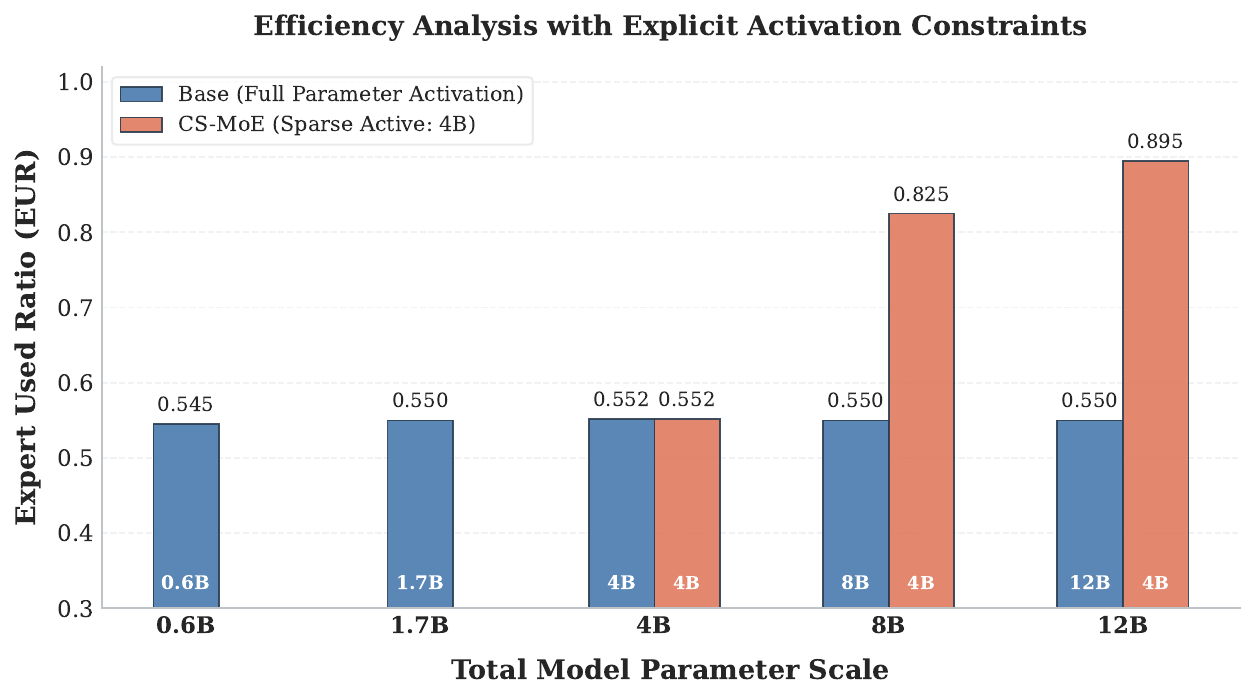}
\caption{Efficiency analysis of CS-MoE models.}
\label{fig:56}
\end{figure}

\section{Experiments}
\subsection{Experimental Setup} 

We evaluate the efficiency, scalability, and structural convergence of \textbf{CS-MoE} across scales ranging from 0.6B to 12B parameters using the Qwen3 architecture~\cite{yang2025qwen3technicalreport} as our experimental backbone. Detailed architectural dimensions and layer allocations are provided in Appendix~\ref{sec:appendix-train}.


\paragraph{Efficiency Metric} 
To quantify parameter reuse within our dual-path design, we define the \textbf{Expert Utilization Ratio (EUR)} ($\rho$):
\[
 \rho = \frac{\left|\bigcup_{l=1}^{L} S_l\right|}{\min(M, \delta)},
\]
where $M$ is the physical capacity of the shared pool, $S_l$ is the set of shared experts activated at layer $l$, and $\delta = \sum_{l=1}^{L} |S_l|$ is the cumulative expert query volume. The term $\min(M, \delta)$ ensures normalization across different allocation budgets.

\subsection{Architectural Efficiency and Scalability}

Our pre-training workflow utilizes an integrated high-quality corpus of \textbf{WuDao}~\cite{yuan2021wudaocorpora} and \textbf{DCLM}~\cite{DataComp-LM} for 12,500 steps with an effective batch size $\approx$ 1M tokens. Perplexity (PPL) is used to trace training dynamics.

\paragraph{Parameter Expansion vs. Dense Baselines}
We compare CS-MoE directly against Dense Transformers under matched parameter and FLOP budgets. As illustrated in Figure~\ref{fig:1}, CS-MoE consistently achieves lower pre-training perplexity across all model scales (exact numerical values in Appendix~\ref{sec:appendix-exact-ppl}, Table~\ref{tab:ppl_figure3}). Crucially, CS-MoE exhibits a strong ``Parameter Expansion'' effect: an 8B CS-MoE model activating only \textbf{55\%} of its physical parameters per forward pass outperforms a fully activated 8B Dense baseline (PPL 9.55 vs.\ 9.75), demonstrating that cross-layer expert reuse expands effective representational capacity without increasing active FLOPs.

\paragraph{Dynamic Compute Allocation}
In Figure~\ref{fig:2}, we evaluate the scaling behavior of CS-MoE by increasing the Top-$K$ activation count while keeping the parameter count fixed. The results demonstrate monotonic perplexity reductions as active compute increases (e.g., 0.6B-A1.7B reaches 13.11 PPL vs.\ 13.48 for 0.6B-A0.6B). Unlike rigid dense models, CS-MoE allows dynamic FLOP scaling at inference time by tuning the routing budget into the centralized pool.

\paragraph{Convergence to Sparse MoE Baselines}
In Figure~\ref{fig:3}, we strictly align both the total parameter footprint (HBM storage) and active FLOPs between CS-MoE and standard MoEs with layer-isolated experts. CS-MoE achieves lower perplexity (9.65 vs.\ 9.70 at 8B scale, 9.19 vs.\ 9.40 at 12B scale). As depicted in Figure~\ref{fig:56}, expanding the shared pool size ($M$) drives the EUR ($\rho$) toward 1.0, asymptotically recovering the capacity upper bound of standard MoEs while maintaining flexible parameter sharing.


\begin{table}[t!]
\centering
\small
\renewcommand{\arraystretch}{1.18}
\caption{\textbf{Downstream benchmark accuracy (\%)} on C-Eval and AQuA across training checkpoints.}
\label{tab:downstream-steps}
\resizebox{\columnwidth}{!}{%
\begin{tabular}{l cccccc}
\toprule
\multirow{2}{*}{\textbf{Steps}} & \multicolumn{2}{c}{\textbf{0.6B / -A0.9B}} & \multicolumn{2}{c}{\textbf{1.7B / -A4B}} & \multicolumn{2}{c}{\textbf{4B / -A8B}} \\
\cmidrule(lr){2-3} \cmidrule(lr){4-5} \cmidrule(lr){6-7}
 & Dense & CS-MoE & Dense & CS-MoE & Dense & CS-MoE \\
\midrule

\rowcolor{headerbg}
\multicolumn{7}{l}{\textbf{C-Eval} \textcolor{subtext}{\small (Multi-discipline Chinese Evaluation)}} \\
2,500  & 28.2 & 31.5 & 30.1 & 31.7 & 29.3 & 29.8 \\
7,500  & 35.1 & 39.3 & 36.4 & 40.1 & 35.6 & 41.0 \\
\rowcolor{defaultbg}
12,500 & 37.5 & \textbf{41.9} & 38.6 & \textbf{42.3} & 40.7 & \textbf{43.2} \\
\midrule

\rowcolor{headerbg}
\multicolumn{7}{l}{\textbf{AQuA} \textcolor{subtext}{\small (Multi-step Algebraic Reasoning)}} \\
2,500  & 22.4 & 26.1 & 21.8 & 28.5 & 27.2 & 33.6 \\
7,500  & 29.1 & 34.0 & 29.8 & 34.3 & 36.1 & 37.2 \\
\rowcolor{defaultbg}
12,500 & 31.2 & \textbf{36.4} & 32.6 & \textbf{37.8} & 39.1 & \textbf{40.5} \\
\bottomrule
\end{tabular}%
}
\end{table}

\subsection{Evaluation on Downstream Benchmarks}
We evaluate cross-scale downstream generalization on \textbf{C-Eval}~\cite{nguyen-etal-2024-ceval-benchmark} (a comprehensive Chinese benchmark spanning 52 domains) and \textbf{AQuA}~\cite{DBLP:conf/acl/LingYDB17} (multi-step algebraic reasoning). As summarized in Table~\ref{tab:downstream-steps}, CS-MoE consistently surpasses the Dense baseline across all evaluated parameter scales and intermediate checkpoints. Crucially, the performance advantage is most pronounced under severe parameter constraints at the 0.6B scale, delivering final gains of +4.4\% on C-Eval and +5.2\% on AQuA; this highlights that concurrent access to a centralized shared pool effectively compensates for the restricted capacity of smaller models. Furthermore, CS-MoE exhibits substantially accelerated convergence dynamics, matching or exceeding the Dense baseline's final performance (12,500 steps) within merely 5,000 to 7,500 training steps, confirming that cross-layer expert sharing expedites knowledge acquisition throughout the pre-training lifecycle.

To evaluate long-term convergence and scaling stability, our 8B CS-MoE model is subjected to full-scale pre-training across a 4.5-trillion-token corpus expanded with \textbf{Ultra FineWeb}~\cite{wang2025ultra} (details in Appendix~\ref{sec:appendix-train}).
The models were evaluated zero/few-shot across seven diverse benchmarks spanning multi-step reasoning (BBH~\cite{suzgun2023challenging}, GSM8K~\cite{cobbe2021training}), domain knowledge (CMMLU~\cite{li2024cmmlu}, C-Eval~\cite{huang2023ceval}, ARC~\cite{clark2018arc}), instruction following (IFEval~\cite{zhou2023ifeval}), and complex mathematics (Competition Math~\cite{hendrycks2021measuring}).

As shown in Table~\ref{tab:4.5t_benchmarks}, CS-MoE achieves substantial improvements over the Dense baseline on complex reasoning (BBH +0.4640), world knowledge (CMMLU +0.0525, ARC +0.0425), and mathematical problem-solving (GSM8K +0.0050). We observed steady convergence with zero expert collapse throughout the 4.5T-token trajectory. Minor trade-offs observed on Competition Math and IFEval can be attributed to raw pre-training alignment sensitivity, which is typically addressed during targeted supervised fine-tuning.

\begin{table}[t!]
\centering
\small
\renewcommand{\arraystretch}{1.15} 
\caption{Downstream performance comparison at 8B scale after full-scale pre-training on 4.5 trillion tokens.}
\label{tab:4.5t_benchmarks}
\resizebox{\columnwidth}{!}{%
\begin{tabular}{lccc}
\toprule
\textbf{Benchmark} & \textbf{8B Dense Baseline} & \textbf{8B-A8B CS-MoE (Ours)} & \textbf{Diff ($\Delta$)} \\
\midrule
\textbf{BBH}~\cite{suzgun2023challenging}     & 0.2867 & 0.7507 & \textcolor{posgreen}{\textbf{+0.4640}} \\
\textbf{CMMLU}~\cite{li2024cmmlu}   & 0.7417 & 0.7942 & \textcolor{posgreen}{\textbf{+0.0525}} \\
\textbf{ARC}~\cite{clark2018arc}     & 0.6875 & 0.7300 & \textcolor{posgreen}{\textbf{+0.0425}} \\
\textbf{GSM8K}~\cite{cobbe2021training}   & 0.9400 & 0.9450 & \textcolor{posgreen}{\textbf{+0.0050}} \\
\textbf{C-Eval}~\cite{huang2023ceval}  & 0.6996 & 0.7038 & \textcolor{posgreen}{\textbf{+0.0042}} \\
\textbf{IFEval}~\cite{zhou2023ifeval}  & 0.4300 & 0.4005 & \textcolor{negred}{-0.0295} \\
\textbf{MATH}~\cite{hendrycks2021measuring}    & 0.7930 & 0.6660 & \textcolor{negred}{-0.1270} \\
\bottomrule
\end{tabular}
}
\end{table}

\subsection{Ablation Studies}
\label{sec:ablations}

\begin{table}[t!]
\centering
\small
\renewcommand{\arraystretch}{1.2} 
\caption{Ablation results of architectural components under matched 4B parameters and 4B activations.}
\label{tab:ablations}
\resizebox{\columnwidth}{!}{%
\begin{tabular}{lccc}
\toprule
\textbf{Ablation Setting} & \textbf{PPL} $\downarrow$ & \textbf{C-Eval} $\uparrow$ & \textbf{AQuA} $\uparrow$ \\
\midrule

\rowcolor{headerbg}
\multicolumn{4}{l}{\textbf{1. Independent Expert Allocation ($N_{indep}$)}} \\
$N_{indep} = 0$ \textcolor{subtext}{\small (Shared pool only)} & 10.32 & 41.8 & 39.5 \\
\rowcolor{defaultbg}
$N_{indep} = 1$ \textbf{(Default)} & \textbf{10.08} & \textbf{43.2} & \textbf{40.5} \\
$N_{indep} = 2$ & 10.11 & 43.0 & 40.3 \\
\midrule

\rowcolor{headerbg}
\multicolumn{4}{l}{\textbf{2. Router Architecture}} \\
Centrally Shared Router & 10.29 & 41.9 & 39.6 \\
\rowcolor{defaultbg}
Layer-Specific Router \textbf{(Default)} & \textbf{10.08} & \textbf{43.2} & \textbf{40.5} \\
\midrule

\rowcolor{headerbg}
\multicolumn{4}{l}{\textbf{3. Load Balancing Coefficient ($\alpha$)}} \\
$\alpha = 0.0$ \textcolor{subtext}{\small (No balancing / Collapse)} & 10.35 & 41.4 & 39.2 \\
\rowcolor{defaultbg}
$\alpha = 0.01$ \textbf{(Default)} & \textbf{10.08} & \textbf{43.2} & \textbf{40.5} \\
$\alpha = 0.1$ & 10.13 & 42.8 & 40.1 \\
$\alpha = 0.5$ & 10.12 & 42.0 & 39.6 \\
\midrule

\rowcolor{headerbg}
\multicolumn{4}{l}{\textbf{4. Cross-Layer Parameter Tying Paradigm}} \\
Static Parameter Tying \textcolor{subtext}{\small (ALBERT-style)} & 12.74 & 38.4 & 37.6 \\
Standard Dense \textcolor{subgray}{\small (Untied Baseline)} & 10.45 & 40.7 & 39.1 \\
\rowcolor{defaultbg}
CS-MoE Dynamic Sharing \textbf{(Ours)} & \textbf{10.08} & \textbf{43.2} & \textbf{40.5} \\

\bottomrule
\end{tabular}%
}
\end{table}
To isolate the contributions of individual architectural choices, we conducted systematic ablation experiments under strictly controlled parameter and activation budgets at the 4B scale, as summarized in Table~\ref{tab:ablations}. 
First, evaluating the independent expert allocation reveals that anchoring at least one layer-private expert ($N_{indep}=1$) is essential for capturing depth-specific syntactic representations; eliminating it entirely ($N_{indep}=0$) causes severe performance degradation (10.32 vs.\ 10.08 PPL), while allocating additional private experts ($N_{indep}=2$) yields diminishing returns due to shrinking the shared pool capacity. 
Second, our layer-specific routing policy markedly outperforms a globally shared router (10.08 vs.\ 10.29 PPL), demonstrating that different network depths require customized gating decisions to process hierarchical abstractions. 
Third, tuning the auxiliary load-balancing coefficient demonstrates that while omitting balancing loss ($\alpha=0$) triggers expert collapse (10.35 PPL), an intermediate penalty ($\alpha=0.01$) effectively balances pool utilization without over-constraining the model's natural routing preferences. 
Finally, benchmarking against classical cross-layer parameter tying illustrates the fundamental advantage of our design: while rigid ALBERT-style weight sharing severely impairs representational capacity (12.74 PPL), CS-MoE's dynamic routing reuses weights adaptively across layers, surpassing even the standard untied Dense baseline (10.45 PPL) while maintaining high parameter efficiency.

\section{Conclusion}

We presented \textbf{CS-MoE}, a Transformer architecture featuring cross-layer expert sharing. By coupling depth-localized independent experts with a globally shared expert pool, CS-MoE breaks inter-layer parameter isolation while preserving hierarchical representation. Extensive evaluations across model scales and training regimes demonstrate its lower perplexity, higher parameter utilization, and superior downstream task performance compared to dense and sparse baselines.

\section*{Limitations}
While CS-MoE demonstrates superior parameter efficiency and promising scaling properties, several limitations and open challenges remain to be addressed in future work:
\begin{itemize}[leftmargin=*]
\item \textbf{Generalization at Ultra-Large Model Scales:} Due to constrained computational and hardware resources, our empirical validation is conducted up to the 12B scale. The scaling dynamics, routing stability, and Pareto efficiency of CS-MoE on ultra-large regimes (e.g., 70B+ parameters) remain to be rigorously verified.

\item \textbf{Distributed Infrastructure and Acceleration:} The architectural topology of CS-MoE fundamentally departs from both dense Transformers and standard layer-isolated MoEs. Current mainstream distributed frameworks (e.g., standard \textit{Expert Parallelism} and \textit{Pipeline Parallelism}) are not tailored for a centralized, cross-layer shared expert pool. While CS-MoE possesses significant untapped potential for training throughput and inference acceleration via cross-layer weight reuse and caching, fully unlocking these speedups requires dedicated system-level infrastructure, such as custom communication primitives and optimized routing kernels.

\item \textbf{Training Paradigms and Adaptation Techniques:} This study strictly evaluates CS-MoE trained under the supervision of next-token prediction. The behavior and efficacy of cross-layer expert sharing under post-training alignment pipelines (e.g., RLHF, DPO, or test-time reinforcement learning) have not yet been established. Additionally, downstream parameter-efficient fine-tuning (PEFT) strategies (e.g., LoRA) warrant specialized study, as naively adapting or tying shared experts across different network depths may require depth-aware adaptation mechanisms.
\end{itemize}

\bibliography{anthology,custom,references}

@article{Yun2024TowardIM,
 author = {Longfei Yun and Yonghao Zhuang and Yao Fu and Eric P. Xing and Hao Zhang},
 booktitle = {arXiv.org},
 journal = {ArXiv},
 title = {Toward Inference-optimal Mixture-of-Expert Large Language Models},
 volume = {abs/2404.02852},
 year = {2024}
}

@article{Shao2024DeepSeekV2AS,
 author = {Zhihong Shao and Damai Dai and Daya Guo and Bo Liu (Benjamin Liu) and Zihan Wang and Huajian Xin},
 booktitle = {arXiv.org},
 journal = {ArXiv},
 title = {DeepSeek-V2: A Strong, Economical, and Efficient Mixture-of-Experts Language Model},
 volume = {abs/2405.04434},
 year = {2024}
}

@article{shazeer2017outrageously,
  title={Outrageously large neural networks: The sparsely-gated mixture-of-experts layer},
  author={Shazeer, Noam and Mirhoseini, Azalia and Maziarz, Krzysztof and Davis, Andy and Le, Quoc and Hinton, Geoffrey and Dean, Jeff},
  journal={arXiv preprint arXiv:1701.06538},
  year={2017}
}

@article{lepikhin2020gshard,
  title={Gshard: Scaling giant models with conditional computation and automatic sharding},
  author={Lepikhin, Dmitry and Lee, Hyoukseok and Xu, Yuanzhong and Chen, Dehao and Firat, Orhan and Huang, Yanping and Krikun, Maxim and Shazeer, Noam and Chen, Zhifeng},
  journal={arXiv preprint arXiv:2006.16668},
  year={2020}
}

@article{fedus2022switch,
  title={Switch transformers: Scaling to trillion parameter models with simple and efficient sparsity},
  author={Fedus, William and Zoph, Barret and Shazeer, Noam},
  journal={Journal of Machine Learning Research},
  volume={23},
  number={120},
  pages={1--39},
  year={2022}
}

@article{dai2024deepseekmoe,
  title={DeepSeekMoE: Towards Ultimate Expert Specialization in Mixture-of-Experts Language Models},
  author={Dai, Damai and Deng, Chengqi and Zhao, Shuo and others},
  journal={arXiv preprint arXiv:2401.06066},
  year={2024}
}

@article{muennighoff2025olmoe,
  title={OLMoE: Open Mixture-of-Experts Language Models},
  author={Muennighoff, Niklas and others},
  journal={arXiv preprint arXiv:2501.00662},
  year={2025}
}

@article{mimo2026flash,
  title={MiMo-V2-Flash: Ultra-Sparse MoE for Agentic Workflows},
  author={Xiaomi AI Team},
  journal={Technical Report},
  year={2026}
}

@inproceedings{sparseswaps2025,
  title={SparseSwaps: Tractable LLM Pruning Mask Refinement at Scale},
  author={OpenReview Contributors},
  booktitle={The Thirteenth International Conference on Learning Representations},
  year={2025}
}

@article{yuan2021wudaocorpora,
  title={WuDaoCorpora: A Super Large-scale Chinese Corpora for Pre-training Language Models},
  author={Yuan, Sha and Zhao, Hanyu and Du, Zhengxiao and Ding, Ning and Liu, Xiao and Cen, Yukuo and Huang, Xu and others},
  journal={AI Open},
  volume={2},
  pages={65--68},
  year={2021},
  publisher={Elsevier}
}

@misc{jiang2024mixtralexperts,
      title={Mixtral of Experts}, 
      author={Albert Q. Jiang and Alexandre Sablayrolles and Antoine Roux and Arthur Mensch and Blanche Savary and Chris Bamford and Devendra Singh Chaplot and Diego de las Casas and Emma Bou Hanna and Florian Bressand and Gianna Lengyel and Guillaume Bour and Guillaume Lample and Lélio Renard Lavaud and Lucile Saulnier and Marie-Anne Lachaux and Pierre Stock and Sandeep Subramanian and Sophia Yang and Szymon Antoniak and Teven Le Scao and Théophile Gervet and Thibaut Lavril and Thomas Wang and Timothée Lacroix and William El Sayed},
      year={2024},
      eprint={2401.04088},
      archivePrefix={arXiv},
      primaryClass={cs.LG},
      url={https://arxiv.org/abs/2401.04088}, 
}

@misc{jin2024moeacceleratingmixtureofexpertsmethods,
      title={MoE++: Accelerating Mixture-of-Experts Methods with Zero-Computation Experts}, 
      author={Peng Jin and Bo Zhu and Li Yuan and Shuicheng Yan},
      year={2024},
      eprint={2410.07348},
      archivePrefix={arXiv},
      primaryClass={cs.LG},
      url={https://arxiv.org/abs/2410.07348}, 
}

@misc{wen2025routeexpertssequencetoken,
      title={Route Experts by Sequence, not by Token}, 
      author={Tiansheng Wen and Yifei Wang and Aosong Feng and Long Ma and Xinyang Liu and Yifan Wang and Lixuan Guo and Bo Chen and Stefanie Jegelka and Chenyu You},
      year={2025},
      eprint={2511.06494},
      archivePrefix={arXiv},
      primaryClass={cs.LG},
      url={https://arxiv.org/abs/2511.06494}, 
}

@inproceedings{yang-etal-2024-xmoe,
    title = "{XM}o{E}: Sparse Models with Fine-grained and Adaptive Expert Selection",
    author = "Yang, Yuanhang  and
      Qi, Shiyi  and
      Gu, Wenchao  and
      Wang, Chaozheng  and
      Gao, Cuiyun  and
      Xu, Zenglin",
    editor = "Ku, Lun-Wei  and
      Martins, Andre  and
      Srikumar, Vivek",
    booktitle = "Findings of the Association for Computational Linguistics: ACL 2024",
    month = aug,
    year = "2024",
    address = "Bangkok, Thailand",
    publisher = "Association for Computational Linguistics",
    url = "https://aclanthology.org/2024.findings-acl.694/",
    doi = "10.18653/v1/2024.findings-acl.694",
    pages = "11664--11674"
}

@inproceedings{press2017using,
  title={Using the Output Embedding to Improve Language Models},
  author={Press, Ofir and Wolf, Lior},
  booktitle={Proceedings of the 15th Conference of the European Chapter of the Association for Computational Linguistics (Volume 2: Short Papers)},
  pages={157--163},
  year={2017},
  publisher={Association for Computational Linguistics}
}

@inproceedings{lan2019albert,
  title={{ALBERT}: A Lite {BERT} for Self-supervised Learning of Language Representations},
  author={Lan, Zhenzhong and Chen, Mingdao and Goodman, Sebastian and Gimpel, Kevin and Sharma, Piyush and Soricut, Radu},
  booktitle={International Conference on Learning Representations},
  year={2020}
}

@inproceedings{he2021deberta,
  title={{DeBERTa}: Decoding-enhanced {BERT} with Disentangled Attention},
  author={He, Pengcheng and Liu, Xiaodong and Gao, Jianfeng and Chen, Weizhu},
  booktitle={International Conference on Learning Representations},
  year={2021}
}

@article{chowdhery2023palm,
  title={{PaLM}: Scaling Language Modeling with Pathways},
  author={Chowdhery, Aakanksha and Narang, Sharan and Devlin, Jacob and Bosma, Maarten and Mishra, Gaurav and Roberts, Adam and Barham, Paul and Chung, Hyung Won and Sutton, Charles and Gehrmann, Sebastian and others},
  journal={Journal of Machine Learning Research},
  volume={24},
  number={240},
  pages={1--113},
  year={2023}
}

@inproceedings{dehghani2018universal,
  title={Universal Transformers},
  author={Dehghani, Mostafa and Gouws, Stephan and Vinyals, Oriol and Uszkoreit, Jakob and Kaiser, {\L}ukasz},
  booktitle={International Conference on Learning Representations},
  year={2019}
}

@inproceedings{nguyen-etal-2024-ceval-benchmark,
    title = "{CE}val: A Benchmark for Evaluating Counterfactual Text Generation",
    author = {Nguyen, Van Bach  and
      Seifert, Christin  and
      Schl{\"o}tterer, J{\"o}rg},
    editor = "Mahamood, Saad  and
      Minh, Nguyen Le  and
      Ippolito, Daphne",
    booktitle = "Proceedings of the 17th International Natural Language Generation Conference",
    month = sep,
    year = "2024",
    address = "Tokyo, Japan",
    publisher = "Association for Computational Linguistics",
    url = "https://aclanthology.org/2024.inlg-main.6/",
    doi = "10.18653/v1/2024.inlg-main.6",
    pages = "55--69"
}

@inproceedings{DBLP:conf/acl/LingYDB17,
  author       = {Wang Ling and
                  Dani Yogatama and
                  Chris Dyer and
                  Phil Blunsom},
  editor       = {Regina Barzilay and
                  Min{-}Yen Kan},
  title        = {Program Induction by Rationale Generation: Learning to Solve and Explain
                  Algebraic Word Problems},
  booktitle    = {Proceedings of the 55th Annual Meeting of the Association for Computational
                  Linguistics, {ACL} 2017, Vancouver, Canada, July 30 - August 4, Volume
                  1: Long Papers},
  pages        = {158--167},
  publisher    = {Association for Computational Linguistics},
  year         = {2017},
  url          = {https://doi.org/10.18653/v1/P17-1015},
  doi          = {10.18653/V1/P17-1015},
  bibsource    = {dblp computer science bibliography, https://dblp.org}
}

@article{wang2025ultra,
  author       = {Yudong Wang and
                  Zixuan Fu and
                  Jie Cai and
                  Peijun Tang and
                  Hongya Lyu and
                  Yewei Fang and
                  Zhi Zheng and
                  Jie Zhou and
                  Guoyang Zeng and
                  Chaojun Xiao and
                  Xu Han and
                  Zhiyuan Liu},
  title        = {Ultra-FineWeb: Efficient Data Filtering and Verification for High-Quality
                  {LLM} Training Data},
  journal      = {CoRR},
  volume       = {abs/2505.05427},
  year         = {2025},
  url          = {https://doi.org/10.48550/arXiv.2505.05427},
  doi          = {10.48550/ARXIV.2505.05427},
  eprinttype   = {arXiv},
  eprint       = {2505.05427}
}

@inproceedings{suzgun2023challenging,
  author       = {Mirac Suzgun and
                  Nathan Scales and
                  Nathanael Sch{\"a}rli and
                  Sebastian Gehrmann and
                  Yi Tay and
                  Hyung Won Chung and
                  Aakanksha Chowdhery and
                  Quoc V. Le and
                  Ed H. Chi and
                  Denny Zhou and
                  Jason Wei},
  editor       = {Anna Rogers and
                  Jordan L. Boyd{-}Graber and
                  Naoaki Okazaki},
  title        = {Challenging {BIG-Bench} Tasks and Whether Chain-of-Thought Can Solve
                  Them},
  booktitle    = {Findings of the Association for Computational Linguistics: {ACL} 2023,
                  Toronto, Canada, July 9-14, 2023},
  series       = {Findings of {ACL}},
  volume       = {{ACL} 2023},
  pages        = {13003--13051},
  publisher    = {Association for Computational Linguistics},
  year         = {2023},
  url          = {https://doi.org/10.18653/v1/2023.findings-acl.824},
  doi          = {10.18653/V1/2023.FINDINGS-ACL.824}
}

@article{cobbe2021training,
  author       = {Karl Cobbe and
                  Vineet Kosaraju and
                  Mohammad Bavarian and
                  Mark Chen and
                  Heewoo Jun and
                  Lukasz Kaiser and
                  Matthias Plappert and
                  Jerry Tworek and
                  Jacob Hilton and
                  Reiichiro Nakano and
                  Christopher Hesse and
                  John Schulman},
  title        = {Training Verifiers to Solve Math Word Problems},
  journal      = {CoRR},
  volume       = {abs/2110.14168},
  year         = {2021},
  url          = {https://arxiv.org/abs/2110.14168},
  eprinttype   = {arXiv},
  eprint       = {2110.14168}
}

@inproceedings{li2024cmmlu,
  author       = {Haonan Li and
                  Yixuan Zhang and
                  Fajri Koto and
                  Yifei Yang and
                  Hai Zhao and
                  Yeyun Gong and
                  Nan Duan and
                  Timothy Baldwin},
  editor       = {Lun{-}Wei Ku and
                  Andre Martins and
                  Vivek Srikumar},
  title        = {{CMMLU:} Measuring massive multitask language understanding in Chinese},
  booktitle    = {Findings of the Association for Computational Linguistics, {ACL} 2024,
                  Bangkok, Thailand and virtual meeting, August 11-16, 2024},
  series       = {Findings of {ACL}},
  volume       = {{ACL} 2024},
  pages        = {11260--11285},
  publisher    = {Association for Computational Linguistics},
  year         = {2024},
  url          = {https://doi.org/10.18653/v1/2024.findings-acl.671},
  doi          = {10.18653/V1/2024.FINDINGS-ACL.671}
}

@inproceedings{huang2023ceval,
  author       = {Yuzhen Huang and
                  Yuzhuo Bai and
                  Zhihao Zhu and
                  Junlei Zhang and
                  Jinghan Zhang and
                  Tangjun Su and
                  Junteng Liu and
                  Chuancheng Lv and
                  Yikai Zhang and
                  Jiayi Lei and
                  Yao Fu and
                  Maosong Sun and
                  Junxian He},
  editor       = {Alice Oh and
                  Tristan Naumann and
                  Amir Globerson and
                  Kate Saenko and
                  Moritz Hardt and
                  Sergey Levine},
  title        = {C-Eval: {A} Multi-Level Multi-Discipline Chinese Evaluation Suite
                  for Foundation Models},
  booktitle    = {Advances in Neural Information Processing Systems 36: Annual Conference
                  on Neural Information Processing Systems 2023, NeurIPS 2023, New Orleans,
                  LA, USA, December 10 - 16, 2023},
  year         = {2023},
  url          = {http://papers.nips.cc/paper_files/paper/2023/hash/c6ec1844bec96d6d32ae95ae694e23d8-Abstract-Datasets_and_Benchmarks.html}
}

@article{clark2018arc,
  author       = {Peter Clark and
                  Isaac Cowhey and
                  Oren Etzioni and
                  Tushar Khot and
                  Ashish Sabharwal and
                  Carissa Schoenick and
                  Oyvind Tafjord},
  title        = {Think you have Solved Question Answering? Try {ARC}, the {AI2} Reasoning
                  Challenge},
  journal      = {CoRR},
  volume       = {abs/1803.05457},
  year         = {2018},
  url          = {http://arxiv.org/abs/1803.05457},
  eprinttype   = {arXiv},
  eprint       = {1803.05457}
}

@article{zhou2023ifeval,
  author       = {Jeffrey Zhou and
                  Tianjian Lu and
                  Swaroop Mishra and
                  Siddhartha Brahma and
                  Sujoy Basu and
                  Yi Luan and
                  Denny Zhou and
                  Le Hou},
  title        = {Instruction-Following Evaluation for Large Language Models},
  journal      = {CoRR},
  volume       = {abs/2311.07911},
  year         = {2023},
  url          = {https://doi.org/10.48550/arXiv.2311.07911},
  doi          = {10.48550/ARXIV.2311.07911},
  eprinttype   = {arXiv},
  eprint       = {2311.07911}
}

@inproceedings{hendrycks2021measuring,
  author       = {Dan Hendrycks and
                  Collin Burns and
                  Saurav Kadavath and
                  Akul Arora and
                  Steven Basart and
                  Eric Tang and
                  Dawn Song and
                  Jacob Steinhardt},
  editor       = {Joaquin Vanschoren and
                  Sai{-}Kit Yeung},
  title        = {Measuring Mathematical Problem Solving With the {MATH} Dataset},
  booktitle    = {Proceedings of the Neural Information Processing Systems Track on
                  Datasets and Benchmarks 1, NeurIPS Datasets and Benchmarks 2021, December
                  2021, virtual},
  year         = {2021},
  url          = {https://datasets-benchmarks-proceedings.neurips.cc/paper/2021/hash/be83ab3ecd0db773eb2dc1b0a17836a1-Abstract-round2.html}
}

@misc{yang2025qwen3technicalreport,
      title={Qwen3 Technical Report}, 
      author={An Yang and Anfeng Li and Baosong Yang and Beichen Zhang and Binyuan Hui and Bo Zheng and Bowen Yu and Chang Gao and Chengen Huang and Chenxu Lv and Chujie Zheng and Dayiheng Liu and Fan Zhou and Fei Huang and Feng Hu and Hao Ge and Haoran Wei and Huan Lin and Jialong Tang and Jian Yang and Jianhong Tu and Jianwei Zhang and Jianxin Yang and Jiaxi Yang and Jing Zhou and Jingren Zhou and Junyang Lin and Kai Dang and Keqin Bao and Kexin Yang and Le Yu and Lianghao Deng and Mei Li and Mingfeng Xue and Mingze Li and Pei Zhang and Peng Wang and Qin Zhu and Rui Men and Ruize Gao and Shixuan Liu and Shuang Luo and Tianhao Li and Tianyi Tang and Wenbiao Yin and Xingzhang Ren and Xinyu Wang and Xinyu Zhang and Xuancheng Ren and Yang Fan and Yang Su and Yichang Zhang and Yinger Zhang and Yu Wan and Yuqiong Liu and Zekun Wang and Zeyu Cui and Zhenru Zhang and Zhipeng Zhou and Zihan Qiu},
      year={2025},
      eprint={2505.09388},
      archivePrefix={arXiv},
      primaryClass={cs.CL},
      url={https://arxiv.org/abs/2505.09388}, 
}

@inproceedings{DataComp-LM,
  author       = {Jeffrey Li and
                  Alex Fang and
                  Georgios Smyrnis and
                  Maor Ivgi and
                  Matt Jordan and
                  Samir Yitzhak Gadre and
                  Hritik Bansal and
                  Etash Kumar Guha and
                  Sedrick Scott Keh and
                  Kushal Arora and
                  Saurabh Garg and
                  Rui Xin and
                  Niklas Muennighoff and
                  Reinhard Heckel and
                  Jean Mercat and
                  Mayee F. Chen and
                  Suchin Gururangan and
                  Mitchell Wortsman and
                  Alon Albalak and
                  Yonatan Bitton and
                  Marianna Nezhurina and
                  Amro Abbas and
                  Cheng{-}Yu Hsieh and
                  Dhruba Ghosh and
                  Josh Gardner and
                  Maciej Kilian and
                  Hanlin Zhang and
                  Rulin Shao and
                  Sarah M. Pratt and
                  Sunny Sanyal and
                  Gabriel Ilharco and
                  Giannis Daras and
                  Kalyani Marathe and
                  Aaron Gokaslan and
                  Jieyu Zhang and
                  Khyathi Raghavi Chandu and
                  Thao Nguyen and
                  Igor Vasiljevic and
                  Sham M. Kakade and
                  Shuran Song and
                  Sujay Sanghavi and
                  Fartash Faghri and
                  Sewoong Oh and
                  Luke Zettlemoyer and
                  Kyle Lo and
                  Alaaeldin El{-}Nouby and
                  Hadi Pouransari and
                  Alexander Toshev and
                  Stephanie Wang and
                  Dirk Groeneveld and
                  Luca Soldaini and
                  Pang Wei Koh and
                  Jenia Jitsev and
                  Thomas Kollar and
                  Alex Dimakis and
                  Yair Carmon and
                  Achal Dave and
                  Ludwig Schmidt and
                  Vaishaal Shankar},
  editor       = {Amir Globersons and
                  Lester Mackey and
                  Danielle Belgrave and
                  Angela Fan and
                  Ulrich Paquet and
                  Jakub M. Tomczak and
                  Cheng Zhang},
  title        = {DataComp-LM: In search of the next generation of training sets for
                  language models},
  booktitle    = {Advances in Neural Information Processing Systems 37: Annual Conference
                  on Neural Information Processing Systems 2024, NeurIPS 2024, Vancouver,
                  BC, Canada, December 10 - 15, 2024},
  year         = {2024},
  url          = {http://papers.nips.cc/paper_files/paper/2024/hash/19e4ea30dded58259665db375885e412-Abstract-Datasets_and_Benchmarks_Track.html}
}

\newpage
\appendix

\section{Related Work}
\label{sec:appendix-related}

\subsection{Mixture-of-Experts Transformers}
Mixture-of-Experts (MoE) has emerged as the leading paradigm for scaling Transformer capacity while maintaining constant computational costs per token. Foundational works established the sparsely-gated MoE layer~\cite{shazeer2017outrageously} and demonstrated its scalability to hundreds of billions of parameters through automatic sharding~\cite{lepikhin2020gshard}. The Switch Transformer~\cite{fedus2022switch} further optimized this via Top-1 routing, significantly improving training efficiency over dense baselines.
Recent innovations have focused on refining routing granularity and expert specialization. DeepSeekMoE~\cite{dai2024deepseekmoe} and DeepSeek-V2~\cite{Shao2024DeepSeekV2AS} introduced fine-grained expert segmentation and shared experts to better capture common knowledge, while Mixtral~\cite{jiang2024mixtralexperts} validated the efficacy of Top-2 routing for open-weight models. Further architectural refinements include OLMoE~\cite{muennighoff2025olmoe}, which provides open-source scaling insights, and MoE++~\cite{jin2024moeacceleratingmixtureofexpertsmethods}, which utilizes zero-computation experts to reduce FLOPs. Efficiency is further pushed by X-MoE~\cite{yang-etal-2024-xmoe} through threshold-based routing and Seqtopk~\cite{wen2025routeexpertssequencetoken}, which shifts from token-level to sequence-level routing to minimize communication overhead. Most recently, MiMo-V2-Flash~\cite{mimo2026flash} has optimized ultra-sparse designs specifically for agentic workflows.
Despite these diversities in routing and granularity, existing MoE architectures share a fundamental constraint: layer-wise isolation. Experts are strictly confined to their respective depths, preventing the exploitation of cross-layer semantic redundancies. Our proposed CS-MoE addresses this limitation by enabling longitudinal expert reusability, thereby maximizing parameter utility across the entire network hierarchy.

\subsection{Parameter Sharing in Transformers}
Parameter sharing has long served as a vital mechanism for enhancing model generalization and efficiency, with early research focusing on weight tying between embedding and output layers~\cite{press2017using} in Transformers. This evolved into the full cross-layer parameter sharing introduced by ALBERT~\cite{lan2019albert}, though its reliance on uniform sharing at every depth inherently limits the model's ability to learn specialized representations. While subsequent investigations in DeBERTa~\cite{he2021deberta} and PaLM~\cite{chowdhery2023palm} identified functional similarities across layers, and Universal Transformers~\cite{dehghani2018universal} proposed recurrent sharing, these approaches often introduce sequential bottlenecks that compromise parallelization. Within the Mixture-of-Experts (MoE) domain, most architectures maintain strict layer-wise isolation of experts~\cite{shazeer2017outrageously, lepikhin2020gshard, fedus2022switch}. DeepSeekMoE~\cite{dai2024deepseekmoe} attempted to bridge this gap with "shared experts," yet these remain restricted to intra-layer application and do not exploit the longitudinal redundancies observed in recent analyses like MoE++~\cite{jin2024moeacceleratingmixtureofexpertsmethods}. CS-MoE addresses these fundamental constraints by implementing a dual-tier hierarchy that balances independent, depth-specific experts with a global shared pool. Unlike the rigid uniformity of ALBERT or the static recurrence of Universal Transformers, CS-MoE utilizes per-token dynamic routing to enable genuine inter-layer allocation, thereby achieving superior parameter efficiency while preserving the essential hierarchical capacity of deep Transformer networks.

\section{Pilot Study}
\label{sec:appendix-pilot}

\subsection{Implementation Details}
We conducted our analysis using a pre-trained Qwen3-MoE-4B model, which features 36 Transformer layers and 4 experts per layer ($N=4$, Top-$K=1$). The study was performed on a held-out validation set of 50,000 tokens sampled from the DCLM corpus to ensure statistical significance in activation patterns.

\textbf{Representational Similarity Analysis}.
To quantify the functional overlap between experts without being misled by the permutation invariance of neural weights, we employed Centered Kernel Alignment (CKA). Unlike Cosine similarity, CKA is invariant to orthogonal transformations and isotropic scaling, making it the standard for comparing internal representations.

For two expert weight matrices $W_i^{(l)}$ and $W_j^{(l')}$ from layers $l$ and $l'$, we computed the linear CKA score as follows:
\[
CKA(K, L) = \frac{HSIC(K, L)}{\sqrt{HSIC(K, K) HSIC(L, L)}}
\]
where $K = XX^T$ and $L = YY^T$ are the Gram matrices of the activations produced by the respective experts, and $HSIC$ is the Hilbert-Schmidt Independence Criterion. A high CKA score ($>0.85$) between a shallow-layer expert and a deep-layer expert served as our primary indicator of longitudinal redundancy.

\textbf{Zero-Shot Expert Substitution Protocol}.
To verify the interchangeability of experts, we implemented a greedy substitution algorithm:
\begin{enumerate}[leftmargin=*]
\item \textbf{Selection:} For each expert in the target ``Deep'' layers (layers 24-32), we identified a ``Source'' expert from the ``Shallow'' layers (layers 1-8) that maximized the CKA score.
\item \textbf{Replacement:} We performed a direct parameter swap, replacing the weights of the Deep expert with the weights of its Shallow counterpart.
\item \textbf{Inference:} We measured the change in model perplexity on the validation set without any gradient updates or fine-tuning.
\item \textbf{Baseline:} As a control, we performed random substitutions of equal magnitude.
\end{enumerate}

The moderate increase in perplexity (under 15 within the acceptable region) during targeted substitution, compared to a sharp divergence in the random baseline, empirically justifies the feasibility of the shared expert pool proposed in CS-MoE.

\begin{table}[t!]
\centering
\small
\caption{Validation perplexity increase ($+\Delta$ PPL) under targeted CKA vs. random expert substitutions.}
\label{tab:bidirectional_cka}
\resizebox{\linewidth}{!}{%
\begin{tabular}{llcc}
\toprule
\textbf{Substitution Type} & \textbf{Direction} & \textbf{+$\Delta$ PPL (ESR=5\%)} & \textbf{+$\Delta$ PPL (ESR=20\%)} \\
\midrule
Targeted (CKA) & Shallow $\to$ Deep & \textbf{+1.22} & \textbf{+3.45} \\
Targeted (CKA) & Deep $\to$ Shallow & +1.54 & +3.98 \\
\midrule
Random Swap    & Shallow $\leftrightarrow$ Deep & +4.20 & +12.80 \\
Random Control & Arbitrary Layer Swap & +4.15 & +15.30 \\
\bottomrule
\end{tabular}
}
\end{table}

\subsection{Extension: Bidirectional CKA and Random Swaps}
To rigorously test cross-layer functional equivalence, we expanded the pilot study on Qwen3-MoE-4B to include bidirectional substitutions (Deep $\to$ Shallow) as well as randomized control swaps under Expert Sharing Ratios (ESR) of 5\% and 20\%.
As shown in Table~\ref{tab:bidirectional_cka}, targeted CKA swaps in both directions incur minimal perplexity increases, while random permutations lead to severe degradation. This demonstrates that functional redundancy is symmetric and structured across network depths.

\begin{table*}[t!]
\centering
\small
\caption{Detailed model hyperparameters. *: Top-$K$ activated experts include independent experts of each layer.}
\label{tab:1}
\resizebox{\linewidth}{!}{%
\begin{tabular}{lcccccccc}
\toprule
Model Type & Total Params & Activated Params & Layers ($L$) & $d_{ffn}$ & $d_{exp}$ & $N_{indep}$ & Shared Pool ($M$) & Top-$K$$^{*}$ \\
\midrule
\textbf{Dense} & 0.6B / 1.7B / 4B / 8B & Same & 16 - 36 & 3,072 - 12,288 & - & - & - & - \\
\textbf{CS-MoE 0.6B} & 0.6B & 0.6B / 0.9B / 1.7B & 28 & 3,072 & 768 & 1 & 84 & 4 / 10 / 21 \\
\textbf{CS-MoE 1.7B} & 1.7B & 1.7B / 4B & 28 & 6,144 & 1,536 & 1 & 84 & 4 / 13 \\
\textbf{CS-MoE 4B} & 4B & 4B / 8B & 36 & 9,728 & 2,432 & 1 & 108 & 4 / 10 \\
\textbf{CS-MoE 8B} & 8B & 4B & 36 & 9,728 & 2,432 & 1 & 324 & 4 \\
\textbf{CS-MoE 8B} & 8B & 8B & 36 & 12,288 & 3,072 & 1 & 324 & 4 \\
\textbf{CS-MoE 12B} & 12B & 4B & 36 & 9,728 & 2,432 & 1 & 540 & 4 \\
\bottomrule
\end{tabular}%
}
\end{table*}

\section{Training Details \& Hyperparameters}
\label{sec:appendix-train}

All standard pre-training runs were conducted using customized Megatron-LM on an $8\times$ NVIDIA H200 GPU node. Hyperparameters are detailed in Table~\ref{tab:hyper}. For the scaling run on the 4.5T-token corpus, gradient accumulation was scaled across 32 nodes with identical learning rate schedules.

Table~\ref{tab:1} details the structural hyperparameters across all model scales evaluated.

\begin{table}[t!]
\centering
\small
\caption{Unified Training Hyperparameters.}
\label{tab:hyper}
\resizebox{\linewidth}{!}{%
\begin{tabular}{ll}
\toprule
\textbf{Hyperparameters} & \textbf{Settings} \\
\midrule
Optimizer & AdamW ($\beta_1=0.9, \beta_2=0.95, \epsilon=1\text{e-}8$) \\
Peak Learning Rate & 3e-4 \\
Weight Decay & 0.01 \\
Batch Size & 512 (effective token count $\approx$ 1M) \\
Warmup Steps & 1,000 \\
Total Steps & 12,500 (Base) / 4.5T tokens (Scaled) \\
LR Schedule & Cosine Annealing to $0.1 \times \text{LR}_{\max}$ \\
Context Length & 2,048 \\
Aux Balancing $\alpha$ & 0.01 \\
\bottomrule
\end{tabular}
}
\end{table}

\section{More Results}
\label{sec:appendix-results}

\subsection{Exact Numerical PPL Tables}
\label{sec:appendix-exact-ppl}
To complement the curves presented in Figures~\ref{fig:1}, \ref{fig:2}, and \ref{fig:3}, we report the exact final pre-training perplexity (PPL) values below.

\begin{table}[t!]
\centering
\small
\caption{Final pre-training perplexity corresponding to Figure~\ref{fig:1} (CS-MoE vs. Dense across scales).}
\label{tab:ppl_figure3}
\resizebox{\linewidth}{!}{%
\begin{tabular}{lccc}
\toprule
\textbf{Model Scale} & \textbf{Dense Baseline} & \textbf{CS-MoE (Ours)} & \textbf{Reduction (\%)} $\downarrow$ \\
\midrule
\textbf{0.6B Scale} & 13.62 & 13.48 & 1.03\% \\
\textbf{1.7B Scale} & 11.30 & 11.12 & 1.59\% \\
\textbf{4B Scale}   & 10.45 & 10.08 & 3.54\% \\
\textbf{8B Scale}   & 9.75  & 9.55  & 2.05\% \\
\bottomrule
\end{tabular}
}
\end{table}

\begin{table}[t!]
\centering
\small
\caption{Final pre-training perplexity corresponding to Figure~\ref{fig:2} (CS-MoE with varying activation counts $K$).}
\label{tab:ppl_figure4}
\resizebox{\linewidth}{!}{%
\begin{tabular}{lcc}
\toprule
\textbf{Model Scale} & \textbf{Activation Configuration} & \textbf{Final Perplexity (PPL)} \\
\midrule
\multirow{3}{*}{\textbf{0.6B Scale}} & 0.6B-A0.6B & 13.48 \\
                                     & 0.6B-A0.9B & 13.18 \\
                                     & 0.6B-A1.7B & 13.11 \\
\midrule
\multirow{2}{*}{\textbf{1.7B Scale}} & 1.7B-A1.7B & 11.12 \\
                                     & 1.7B-A4B   & 10.82 \\
\midrule
\multirow{2}{*}{\textbf{4B Scale}}   & 4B-A4B     & 10.08 \\
                                     & 4B-A8B     & 9.95  \\
\bottomrule
\end{tabular}
}
\end{table}

\begin{table}[t!]
\centering
\small
\caption{Final pre-training perplexity corresponding to Figure~\ref{fig:3} (CS-MoE vs. Standard MoE under double-aligned total and active parameters).}
\label{tab:ppl_figure5}
\resizebox{\linewidth}{!}{%
\begin{tabular}{lccc}
\toprule
\textbf{Configuration} & \textbf{Standard MoE} & \textbf{CS-MoE (Ours)} & \textbf{Reduction (\%)} $\downarrow$ \\
\midrule
\textbf{8B-A4B}  & 9.70 & 9.65 & 0.52\% \\
\textbf{12B-A4B} & 9.40 & 9.19 & 2.23\% \\
\bottomrule
\end{tabular}
}
\end{table}

\subsection{Statistical Significance and Seed Variance}
\label{sec:appendix-variance}

To ensure observed performance gains are statistically robust, we conducted 3 independent training runs with distinct random seeds at the 4B scale (12.5k steps). 
As shown in Table~\ref{tab:significance}, CS-MoE's performance gains consistently exceed the standard deviations across all evaluation metrics.

\begin{table}[t!]
\centering
\small
\caption{Mean and standard deviation over 3 independent training runs at 4B scale.}
\label{tab:significance}
\resizebox{\linewidth}{!}{%
\begin{tabular}{lccc}
\toprule
\textbf{Model} & \textbf{Pre-train PPL} & \textbf{C-Eval (\%)} & \textbf{AQuA (\%)} \\
\midrule
4B Dense Baseline & 10.25 $\pm$ 0.04 & 40.8 $\pm$ 0.3 & 39.0 $\pm$ 0.5 \\
4B CS-MoE (Ours)    & \textbf{10.06 $\pm$ 0.03} & \textbf{42.3 $\pm$ 0.2} & \textbf{40.5 $\pm$ 0.4} \\
\bottomrule
\end{tabular}
}
\end{table}

\subsection{Hardware Profiling \& Latency Analysis}
\label{sec:appendix-hardware}

We profiled the computational and memory overhead on an $8\times$ NVIDIA H200 (141GB HBM3e) GPU cluster. 

\paragraph{Routing Computational Cost}
For hidden dimension $d=4096$ and shared pool $M=324$, the gating projection matrix has $4096 \times 324 \approx 1.3\text{M}$ parameters. Computed at each layer, gating operations account for less than 0.1\% of the total FLOPs per forward pass.

\paragraph{Inference Latency with Expert Caching}
Because adjacent layers frequently access overlapping subsets of shared experts, active weights can be retained in GPU L2/SRAM cache across sequential layer executions. Table~\ref{tab:latency} reports latency measurements at batch size 1.

\begin{table}[t!]
\centering
\small
\caption{Real-world latency profiling at 8B model scale on $8\times$ NVIDIA H200 GPUs.}
\label{tab:latency}
\resizebox{\linewidth}{!}{%
\begin{tabular}{lcc}
\toprule
\textbf{Model Configuration} & \textbf{Prefill (ms)} & \textbf{Decoding (ms/tok)} \\
\midrule
Dense 8B                          & 12.5 & 14.2 \\
Standard Sparse MoE 8B            & 14.1 & 16.5 \\
CS-MoE 8B (Default)                 & 13.8 & 15.8 \\
CS-MoE 8B (with Cache Optimization) & \textbf{13.1} & \textbf{14.9} \\
\bottomrule
\end{tabular}
}
\end{table}

With cache optimization, CS-MoE approaches the inference speed of Dense models and outperforms standard MoEs due to a 45\% reduction in unique memory traffic per token.

\subsection{Expert Routing Dynamics across Layers}
\label{sec:appendix-routing-dynamics}

We analyzed the routing patterns across all $L=36$ layers of the fully-trained 8B CS-MoE model:
\begin{itemize}[leftmargin=*]
    \item \textbf{Layer-wise Active Overlap:} The overlap ratio of the top-10\% most frequently activated shared experts between adjacent layers ($l$ and $l+1$) is \textbf{68.4\%}, gradually decaying to \textbf{42.1\%} between distant layers ($l$ and $l+5$). This confirms smooth longitudinal semantic transitions.
    \item \textbf{Expert Specialization (Gini Index):} The average Gini coefficient of expert activation frequencies across layers is \textbf{0.28} (where 0 indicates perfectly uniform global usage and 1 indicates single-layer isolation). This confirms that shared experts are globally utilized rather than confined to specific depths.
\end{itemize}

\end{document}